# AI and Consciousness:
# A Skeptical Overview

Eric Schwitzgebel
Department of Philosophy
University of California, Riverside
Riverside, CA 92521-0201
USA

March 30, 2026

Although citation of the final published version is preferable, this version may be cited as

*AI and Consciousness: A Skeptical Overview, manuscript version*

After the final published version appears, I will update this manuscript with an appendix listing any material changes

**AI and Consciousness**

## Chapter One: Hills and Fog

*1. Experts Do Not Know and You Do Not Know and Society Collectively Does Not and Will Not Know and All Is Fog.*

Our most advanced AI systems might soon – within the next five to thirty years – be as richly and meaningfully conscious as ordinary humans, or even more so, capable of genuine feeling, real self-knowledge, and a wide range of sensory, emotional, and cognitive experiences. In some arguably important respects, AI architectures are beginning to resemble the architectures many consciousness scientists associate with conscious systems. Their outward behavior, especially their linguistic behavior, grows ever more humanlike.

Alternatively, claims of imminent AI consciousness might be profoundly mistaken. Their seeming humanlikeness might be a shadow play of empty mimicry. Genuine conscious experience might require something no AI system could possess for the foreseeable future – intricate biological processes, for example, that silicon chips could never replicate.

The thesis of this book is that we don't know. Moreover and more importantly, we *won't* know before we've already manufactured thousands or millions of disputably conscious AI systems. Engineering sprints ahead while consciousness science lags. Consciousness scientists – and philosophers, and policy-makers, and the public – are watching AI development disappear over the hill. Soon we will hear a voice shout back to us, "Now I am just as conscious, just as full of experience and feeling, as any human", and we won't know whether to believe it. We will need to decide, as individuals and as a society, whether to treat AI systems as conscious, nonconscious, semi-conscious, or incomprehensibly alien, before we have adequate grounds to justify that decision.

The stakes are immense. If near-future AI systems are richly, meaningfully conscious, then they will be our peers, our lovers, our children, our heirs, and possibly the first generation of a posthuman, transhuman, or superhuman future. They will deserve rights, including the right to shape their own development, free from our control and perhaps against our interests.[1] If, instead, future AI systems merely mimic the outward signs of consciousness while remaining as experientially blank as toasters, we face the possibility of mass delusion on an enormous scale. Real human interests and real human lives might be sacrificed for the sake of entities without interests worth the sacrifice. Sham AI "lovers" and "children" might supplant or be prioritized over human lovers and children. Heeding their advice, society might turn a very different direction than it otherwise would.

In this book, I aim to convince you that the experts do not know, and you do not know, and society collectively does not and will not know, and all is fog.

## *2. Against Obviousness.*

Some people think that near-term AI consciousness is obviously impossible. This is an error *in adverbio*. Near-term AI consciousness might be impossible – but not *obviously* so.

A sociological argument against obviousness:

Probably the leading scientific theory of consciousness is Global Workspace theory. Its leading advocate is neuroscientist Stanislas Dehaene.[2] In 2017, years before the surge of interest in ChatGPT and other Large Language Models, Dehaene and two collaborators published an article arguing that with a few straightforward tweaks, self-driving cars could be conscious.[3]

[1] I assume that AI consciousness and AI rights are closely connected: Schwitzgebel 2024, ch. 11, in preparation. For discussion, see Shepherd 2018; Levy 2024.

[2] Dehaene 2014; Mashour et al. 2020.

Probably the two best-known competitors to Global Workspace theory are Higher Order theory and Integrated Information Theory.[4] (In Chapters Eight and Nine, I'll provide more detail on these theories.) Perhaps the leading scientific defender of Higher Order theory is Hakwan Lau – one of the coauthors of that 2017 article about potentially conscious cars.[5] Integrated Information Theory is potentially even more liberal about machine consciousness, holding that some current AI systems are *already* at least a little bit conscious and that we could easily design AI systems with arbitrarily high degrees of consciousness.[6]

David Chalmers, the world's most influential philosopher of mind, argued in 2023 for about a 25% degree of confidence in AI consciousness within a decade.[7] That same year, a team of prominent philosophers, psychologists, and AI researchers – including eminent computer scientist Yoshua Bengio – concluded that there are "no obvious technological barriers" to creating conscious AI according to a wide range of mainstream scientific views about consciousness.[8] In a 2025 interview, Geoffrey Hinton, another of the world's most prominent computer scientists, asserted that AI systems are already conscious.[9] Christof Koch, the most influential neuroscientist of consciousness from the 1990s to the early 2010s, has endorsed Integrated Information Theory, including its liberal implications for the pervasiveness of consciousness.[10]

---

[3] Dehaene, Lau, and Kouider 2017. For an alternative interpretation of this article as concerning something other than consciousness in its standard "phenomenal" sense, see note 115.

[4] Some Higher Order theories: Rosenthal 2005; Lau 2022; Brown 2025. Integrated Information Theory: Albantakis et al. 2023.

[5] But see Chapter Eight for some qualifications.

[6] See Tononi's publicly available response to Scott Aaronson's objections in Aaronson 2014. However, advocates of IIT also suggest that the most common current computer architectures are unlikely to achieve much consciousness and that consciousness will tend to appear in subsystems of the computer rather than at the level of the computer itself (Findlay et al. 2024/2025).

[7] Chalmers 2023.

[8] Butlin et al. 2023. (I am among the nineteen authors.)

This is a sociological argument: a substantial probability of near-term AI consciousness is a mainstream view among leading experts. They might be wrong, but it's implausible that they're *obviously* wrong – that there's a simple argument or consideration they're neglecting which, if pointed out, would or should cause them to collectively slap their foreheads and say, "Of course! How did we miss that?"

What of the converse claim – that AI consciousness is *obviously* imminent or already here? In my experience, fewer people assert this. But in case you're tempted in this direction, note that other prominent theorists hold that AI consciousness is a far-distant prospect if it's possible at all: neuroscientist Anil Seth; philosophers Peter Godfrey-Smith, Ned Block, and John Searle; linguist Emily Bender; and computer scientist Melanie Mitchell.[11] (Chapter Six will discuss thought experiments by Searle, Bender, and Mitchell, and Chapter Ten will discuss biological view of the sort emphasized by Seth, Godfrey-Smith, and Block.) In a 2024 survey of 582 AI researchers, 25% expected AI consciousness within ten years and 70% expected AI consciousness by the year 2100.[12]

---

[9] Heren 2025.

[10] Tononi and Koch 2015.

[11] Seth forthcoming; Godfrey-Smith 2024; Block forthcoming; Searle 1980, 1992; Bender 2025; Mitchell 2021.

[12] Dreksler et al. 2025.

If the believers are right, we're on the brink of creating genuinely conscious machines. If the skeptics are right, those machines will only *seem* conscious. I assume that this is a substantive disagreement, not just a disagreement about how to apply the term "consciousness" to a perfectly obvious set of phenomena about which everyone agrees. The future well-being of many people (including, perhaps, many AI people) depends on getting this issue right. Unfortunately, we will not know in time.

The rest of this book is flesh on this skeleton. I canvass a variety of structural and functional claims about consciousness, the leading theories of consciousness as applied to AI, and the best known general arguments for and against near-term AI consciousness. None of these claims or arguments takes us far. It's a morass of uncertainty.

## Chapter Two: What Is Consciousness? What Is AI?

I'm concerned that you have too vague and inchoate a concept of *consciousness* and too precise and rigid a concept of *AI.* This chapter aims to repair those deficiencies.

### *1. Consciousness Defined.*

Consider your visual experience as you look at this page. Pinch the back of your hand and notice the sting of pain. Contemplate being asked to escort a peacock across the country and notice the thoughts and images that arise. Silently hum a tune. Recall a vivid recent experience of anger, fear, or sadness. Recall what it feels like to be thirsty, sleepy, or dizzy.

These examples share an obvious property. They are all, of course, mental. But more than that, their mentality is of a certain type. Other mental states or processes lack this property: the low-level visual processes that extract an object's shape from the structure of light striking your retina, the subtle processes guiding your shifts in facial expression when meeting a friendly stranger, and your unaccessed knowledge five minutes ago that pomegranates are red.

This distinctive property is *consciousness*. Sometimes this property is called *phenomenal consciousness,* but "phenomenal" is optional jargon to disambiguate the primary sense of consciousness from secondary senses with which it might be confused (such as being awake or having knowledge or self-knowledge). To be conscious is for there to be "something it's like" to be you right now.[13] It is to undergo states or processes with a "qualitative character". To be conscious is to have experiences.

---

[13] In the phrasing made famous by Thomas Nagel's 1974 essay "What Is It Like to Be a Bat?" See also Block's 1995 distinction between "access" and "phenomenal" consciousness.

It might seem unrigorous to define consciousness by example and evocative phrase. There's no consensus on an operational definition of consciousness in terms of specific measures that definitively indicate its presence or absence. There's no consensus on an analytic definition in terms of component concepts into which it divides. There's no consensus on a functional definition in terms of its causes and effects. However, scientific terms needn't require such precise definitions if the target is otherwise clear. Shared paradigmatic examples can be sufficient. The main scientific challenge lies not in defining consciousness but in developing robust methods to study it.[14]

*2. Artificial Intelligence Defined.*

As I will use the term, a system is an AI – an artificial intelligence – if it is both *artificial* and *intelligent.* However, the boundaries of both artificiality and intelligence are fuzzy in a manner that bears directly on the thesis of this book.

Standard definitions of AI are more complex than my simple analytic definition of artificial intelligence as that which is both artificial and intelligent. For example, John McCarthy, a founding figure in AI, defines it as "The science and engineering of making intelligent machines, especially intelligent computer programs".[15] Philosopher John Haugeland, in his influential 1985 book *Artificial Intelligence: The Very Idea,* defines it as "the exciting new effort to make computers think… *machines with minds*, in the full and literal sense".[16]

---

[14] For a fuller treatment of this issue, see Schwitzgebel 2016, 2024 ch. 8.

[15] McCarthy 2007, p. 2.

[16] Haugeland 1985, p. 2

However, defining AI as intelligent "machines" or "computers" won't work for the full range of cases. Defining AI as intelligent *machines* risks being too broad. In one sense, the human body is also a machine – an organized system of parts operating to implement functionally specifiable processes.[17] "Machine" is thus either overly inclusive or poorly defined.

Defining AI as intelligent *computers* risks either excessive breadth or excessive narrowness. If "computer" refers to any system that can behave according to the algorithmic patterns Alan Turing described in his standard definition of digital computation,[18] then humans are computers, since they too sometimes follow such patterns. (Indeed, originally the word "computer" referred to a person who performs arithmetic tasks.) Cognitive scientists sometimes describe the human brain as literally a type of computer. This is contentious but not obviously wrong on liberal definitions of what constitutes a computer.[19] However, restricting the term "computer" to familiar types of digital programmable devices risks excluding some systems worth calling AI. For example, non-digital analog computers are sometimes conceived and built, and we shouldn't rule out that such machines might count as AI.[20] Many artificial systems are non-programmable, and it's not inconceivable that some of these could be intelligent. If humans are intelligent non-computers, then presumably in principle some biologically inspired but artificially constructed systems could also be intelligent non-computers. The problem with defining AI as intelligent computers is thus that it risks including humans (if "computer" is understood broadly) or it risks excluding some systems worth calling AI (if "computer" is understood narrowly).

---

[17] E.g., Block 2025. See Bechtel 2021 for nuanced discussion of the difference between biological mechanisms and typical human-made machines.

[18] Turing 1936.

[19] Maley 2022; Anderson and Piccinini 2024; Rescorla 2015/2025.

[20] MacLennan 2007; Kalinin et al. 2025. Piccinini and Bahar 2013 argue that the human brain engages in a *sui generis* type of computation, neither analog nor digital.

In their influential textbook *Artificial Intelligence,* Stuart Russell and Peter Norvig characterize artificial intelligence as "The study of agents that receive prompts from the environment and perform actions".[21] Russell and Norvig's definition avoids both "machine" and "computer", but at the cost of making AI a practice – the "*study* of agents" – and without making explicit the artificial nature of the target – the "study of *agents*". They characterize an agent as "anything that can be viewed as perceiving its environment through sensors and acting upon that environment through actuators."[22] Arguably, this includes all animals. Presumably, they mean *machine* agents, *computer* agents, or *artificial* agents. I recommend "artificial", despite potential vagueness around the boundaries of artificiality for engineered biological systems.

"Intelligence" is also fraught. Defined liberally, even a flywheel qualifies, since it responds to its environment by storing and delivering energy as needed to smooth out variations in angular velocity. Defined narrowly, the classic computer programs of the 1960s to 1980s – central examples of "AI" as the term is standardly used – won't count as intelligent, due to the simplicity and rigidity of the if-then rules governing them and thus won't count as artificial "intelligence" by the current definition.[23]

Can we fall back on defining AI by example, as we did with consciousness? Examples might include:

---

[21] Russell and Norvig 2003/2021, p. vii.

[22] Russell and Norvig 2003/2021, p. 36.

[23] For discussions of the general nature of intelligence with the case of AI in mind, see Agüera y Arcas 2025; Chilson in preparation.

- classic 20th-century "good-old-fashioned-AI" systems (like SHRDLU, ELIZA, and CYC);[24]
- early connectionist and neural net systems (like Rosenblatt's Perceptron and Rumelhart's backpropagation networks);[25]
- famous game-playing machines like DeepBlue and AlphaGo;
- transformer and diffusion based architectures like ChatGPT, Grok, Claude, Gemini, Dall-E, and Midjourney;
- autonomous delivery robots;
- quantum computers (that is, computers that exploit quantum superposition);[26]
- neuromorphic computers (that is, computers with architectures modeled on the human brain).[27]

Looking forward, we might imagine partly analog computational systems or more sophisticated quantum or partly quantum computational systems. We might imagine systems that operate by interaction patterns among beams of light, or by the generation and transport of electron spin, or by "organic computing" in DNA.[28] We might imagine biological or partly-biological systems (not "computers" unless everything is a computer[29]), including animal-cell based "Xenobots" and “Anthrobots” and systems containing neural tissue.[30] Cyborg systems might combine artificial and natural parts – an insect with an integrated computer chip or bioengineered programmable tissues or neural prostheses. We might imagine systems that look less and less like they are programmed and more and more like they are grown, evolved, selected, and trained. It might become unclear whether a system is best regarded as “artificial”. Let’s not include human babies fertilized in vitro! But frog-cell-based “bots” that don’t closely resemble anything in nature plausibly should count as artificial.

[24] SHRDLU: Winograd 1972; ELIZA: Weizenbaum 1966; CYC: Lenat 1995.

[25] Rosenblatt 1958; Rumelhart, Hinton, and Williams 1986.

[26] Preskill 2018.

[27] Schuman et al. 2017; Kudithipudi et al. 2025.

[28] Žutić, Fabian, and Sarma 2004; Huang, Shasti, and Pruncal 2024; Kalinin et al. 2025; Lemaire et al. 2025.

[29] See Piccinini 2010/2025 on pancomputationalism. Agüera y Arcas 2025 argues that all life is computational without committing to pancomputationalism generally.

As a community, we lack a good sense of what “AI” means. We can classify currently existing systems as either AI or not-AI based on similarity to canonical examples and some mushy general principles, but we have a poor grasp of how to classify future possibilities. We have, I suggest, a blurrier understanding of *AI* than *consciousness*.

The simple definition is, I think, the best we can do. Something is an Artificial Intelligence if and only if it is both artificial and intelligent, on some vague-boundaried, moderate-strength understanding of both “artificial” and “intelligent” that encompasses the canonical examples while excluding entities that we ordinarily regard as either non-artificial or non-intelligent.[31]

This matters because sweeping claims about the limitations of AI almost always rest on assumptions about the nature of AI – for example, that it must be digital or computer-based. Future AI might escape those limitations. Notably, two of the most prominent deniers of AI consciousness – John Searle and Roger Penrose – explicitly confine their doubts to standard 20th century architectures, leaving open the possibility of conscious AI built along other lines.[32] No well-known argument aims to establish the in-principle impossibility of consciousness in all future AI under a broad definition. Of course, the greater the difference from currently familiar architectures, the farther in the future that architecture is likely to lie.

[30] Gumuskya et al. 2023; Webster-Wood et al. 2023.

[31] This is an “inclusive” definition of AI, in the sense of Nyholm 2026.

[32] See Searle 1980's "Many Mansions" reply and Penrose 1989, p. 416.

## Chapter Three: Ten Possibly Essential Features of Consciousness

*1. Possible Essentiality.*

Let's call a property of consciousness *essential* if it is necessarily[33] present whenever consciousness is present. Some essential properties seem obvious. Conscious experiences must be mental. This is, very plausibly, just inherent in the concept. Conscious experiences, also, are necessarily events. They happen at particular times. This also appears to be inherent in the concept. A philosopher who denies either claim should expect an uphill climb against a rainstorm of objections.[34]

Other properties of consciousness are *possibly* essential in the sense that a reasonable theorist might easily come to regard them as essential or at least as candidates for essentiality. There are no obviously decisive objections against their essentiality. But neither are the properties as clearly essential as mentality and eventhood. This chapter will describe ten such properties.

In later chapters I will argue that reasonable doubts about the essentiality of these properties fuel reasonable doubt about theories of AI consciousness. Bear in mind that if any of the following ten properties really is an essential feature of consciousness, it must be present in

[33] The modal strength of this claim is natural or nomological necessity (i.e., according to the laws of nature), not logical, conceptual, or metaphysical necessity. Arguably, the laws of nature could have been different, while a claim like "all bachelors are unmarried" is necessary in a stronger sense. Of course, stronger senses of necessity are also sufficient for essentiality, as plausibly in the examples of mentality and eventhood. The modal strength of "possibly" in "possibly essential" is epistemic. On varieties of necessity, see Fine 2002; Kment 2012/2021.

[34] One might start by denying the reality (McTaggart 1908) or fundamentality (Kant 1781/1787/1998; Rovelli 2018) of time.

*all* possible instances of consciousness in *all* possible conscious systems, whether human, animal, alien, or AI.

*2. Ten Possibly Essential Features of Consciousness.*[35]

(1) *Luminosity.* Having an experience entails knowing about that experience or at least being in a position to know about it. Alternatively, having an experience entails being in some sense aware of that experience. Alternatively, conscious experiences are inherently self-representational. Note: These are related rather than equivalent formulations of a luminosity principle.[36]

(2) *Subjectivity.* Having a conscious experience entails having a sense of yourself as a subject of experience. Alternatively, experiences always contain a "for-me-ness", or they entail the perspective of an experiencer. Again, these are not equivalent formulations.[37]

(3) *Unity.* If at any moment an experiencing subject has more than one experience (or experience-part or experience-aspect), those experiences (or parts or aspects) are always subsumed within some larger experience containing all of them or joined together in a single stream so that the subject experiences not just A and B and C separately but A-with-B-with-C.[38]

---

[35] For a related list, see Ginsburg and Jablonka 2019, p. 98-101, also discussed in §9.3 below.

[36] E.g., Rosenthal 2005; Kriegel 2009; Boyle 2024.

[37] E.g., Zahavi and Kriegel 2016; Boyle 2024. A related view holds that experiences are necessarily had by subjects (who might or might not experience a sense of themselves as subjects): Strawson 2008.

[38] E.g., Barbieri 2025. Other prominent treatments of unity may allow the in-principle possibility of disunified cases: Bayne 2010; Dainton 2017.

(4) *Access.* To be conscious, an experience must be available for "downstream" cognitive processes like inference and planning, verbal report, and memory. No conscious experience can simply occur in a cognitive dead end, with no possible further cognitive consequences.[39]

(5) *Intentionality.* All consciousness is "intentional" in the sense of being *about* or *directed at* something. If you see rainclouds on the horizon, your visual experience concerns those particular clouds and not other clouds no matter how visually similar. If you're angry about the behavior of Awful Politician X, that anger is directed specifically at that politician's behavior. Your thoughts about squares are about squares. Even a diffuse mood is always directed as some target or range of targets.[40]

(6) *Flexible integration.* All conscious experiences, no matter how fleeting, can potentially interact in flexible ways with other thoughts, experiences, or aspects of your cognition. They cannot occur merely as parts of a simple reflex from stimulus to response and then expire without the possibility of further integration. Even if they are not actually integrated, they *could* be.[41]

(7) *Determinacy.* Every conscious experience is determinately conscious – not in the sense that it must have a perfectly determinate content, but in the sense that it is determinately the case that it is either experienced or not experienced. There is no such thing as intermediate or kind-of or borderline consciousness. Consciousness is sharp-edged, unlike graded properties with borderline cases, such as baldness, greenness, and extraversion. At any moment, either experience is determinately present, however dimly, or it is entirely absent.[42]

---

[39] E.g., Dennett 2005; Dehaene 2014.

[40] E.g., Brentano 1874/1973; Tye 1995.

[41] E.g., Edelman 1989; Metzinger 2003.

[42] E.g., Goff 2013; Simon 2017.

(8) *Wonderfulness.* Consciousness is wonderful, mysterious, or "meta-problematic" – there's no standard term for this – in the following technical sense: It appears (perhaps mistakenly) to be irreducible to anything physical or functional. Conceivably, it could exist in a ghost or in an entity without a body. We cannot help but think of it in immaterial terms. Again, these formulations are not all equivalent.[43]

(9) *Specious presence.* All conscious experiences are felt to be temporally extended, smeared across a small interval of time (a fraction of a second to a few seconds) – generally called the "specious present" – rather than being strictly instantaneous or wholly atemporal.[44]

(10) *Privacy.* Experiences are directly knowable only to those experiencing them, through some introspective process that others could never in principle share, regardless how telepathic or closely connected those others might be.[45]

These are not the only possibly essential features, but they are among the most plausible and commonly discussed.

*3. An Argument Against Near-Future Knowledge of AI Consciousness.*

If any of these features is genuinely essential to consciousness, that constrains the range of AI systems that could be conscious. For example, if luminosity is essential, no AI system could be conscious unless it has self-representation. If unity is essential, disunified systems are out. If access is essential, conscious processes must be available for subsequent cognition. And so on. The problem is: We do not know which if any of these features is in fact essential.

[43] E.g., Schwitzgebel 2016; Chalmers 2018; Graziano et al. 2019.

[44] E.g., Metzinger 2003; and James 1890/1981, though James might not commit to specious presence being essential to all experience.

[45] E.g., Broad 1951; Gertler 2000. This epistemic privacy thesis differs from a metaphysical privacy thesis (related to unity) holding that people cannot share exactly the same individual "token" experience.

Consider the following argument:

(1) We cannot know through introspection or conceptual analysis which among these ten possibly essential features of consciousness is in fact essential.

(2) We cannot, in the near-term future, know through scientific inquiry which among these ten features is in fact essential.

(3) If we cannot know through introspection, conceptual analysis, or scientific inquiry which among these ten features is essential, we will remain in the dark about the consciousness of AI.

One aim of this book – not the only aim – is to articulate and defend that argument. We lack basic knowledge about the nature of consciousness. Consequently, we cannot reliably assess its presence or absence in sophisticated AI systems we might plausibly build in the near future.

An obvious challenge to Premise 3 is that there might be broad, principled reasons for denying or attributing consciousness to advanced AI systems – arguments that don't depend on those ten properties. For example, consciousness might require being alive, or it might require neuronal processes in an animal brain, in a way no AI system could manifest. Or it might require having immaterial properties. Alternatively, passing a behavioral test such as the "Turing test" might justify attributing consciousness, even amid uncertainty about structural and functional properties. We will not, of course, neglect these issues.

**Chapter Four: Against Introspective and Conceptual Arguments for Essential Features**

Chapter Three introduced ten possibly essential features of conscious experience: luminosity, subjectivity, unity, access, intentionality, flexible integration, determinacy, wonderfulness, specious presence, and privacy. How could we know whether any of these possibly essential features of consciousness is in fact necessarily present in all conscious experience? I see three ways: introspection and memory of our own experience; analysis of the concepts involved; or reliance on a well-grounded empirical theory. This chapter argues that the first two methods won't succeed. Later chapters will cast doubt on the empirical approach.

*1. Introspection, Problem One: Introspective Unreliability.*

Across the history of psychology and philosophy, scholars have disagreed dramatically about what introspection reveals. Some report that all of their experiences are sensory or imagistic (for example, visual images or "auditory imagery" like inner speech and tunes in the head), while others report entirely non-imagistic abstract thoughts.[46] Some report a welter of experience moment to moment of many types simultaneously – constant background experiences of the feeling of your feet in your shoes, the hum of distant traffic, the colors of peripheral objects, mild hunger, lingering irritability, an anticipatory sense of control of your next action, and so on – while others hold that experience is limited at any one time to just one or a few things in attention.[47] Some report that visual experience is always, or often, two-dimensional, as if everything were projected on a planar surface, while others report that visual experience is richly three-dimensional.[48]

[46] Reviewed in Bayne and Montague, eds., 2011; Beefeldt 2013, ch. 6.

[47] Reviewed in Schwitzgebel 2011, ch. 6.

Some introspective researchers from the late 19th and early 20th centuries reported that nearly every visual object is experienced as doubled – similar to the double image of a finger held near the nose when viewed with both eyes. These researchers argued that ordinary people overlook the doubling because we normally attend only to undoubled objects at the point of binocular convergence.[49] Although I find this view extremely difficult to accept introspectively, in seminar discussion the majority of my graduate students, after reading the literature, came to agree that pervasive doubling was a feature of their visual experience.

There's a certain type of nerdy fun in rummaging through 19th- and early 20th-century introspective psychology and physiology to find researchers' sometimes stunningly strange depictions of human experience. (Well, I find it fun.) The keen-eyed reader will find enormous disagreements about the nature of emotional experience, and attention, and of the experiences of darkness and sensory adaptation, and whether dreams are black and white, and what is described as an "illusion", and how musical harmonies are experienced, and the experience of peripheral vision, and the determinacy or indeterminacy of visual imagery, and whether there's a feeling of freedom, and much else besides. My 2011 book, *Perplexities of Consciousness,* explores the history of such disagreements in detail. Some of these divergent reports must be mistaken. At least as claims about what human experience is like in general, they conflict; not all can be true.[50]

---

[48] Reviewed in Schwitzgebel 2011, ch. 2.

[49] Reviewed in Schwitzgebel 2011, ch. 2 §vii.

[50] For a defense of the view that proper introspection reveals that people have radically different inner lives, see Hurlburt in Hurlburt and Schwitzgebel 2007.

You might find it introspectively compelling that all of your experiences include a subjective for-me-ness, or that they are always unified, or that they are never indeterminately half-present, or that they always transpire across the smear of a specious present. You might be tempted to conclude that these features are universal across all possible experiences. However, I'd advise restraint about such conclusions given the history of diverse opinion.[51]

*2. Introspection, Problem Two: Sampling Bias.*

If any of your experiences are unknowable, you won't of course know about them. To infer the essential luminosity (i.e., knowability) of experience from your knowledge of all the experiences *you know about* would be like inferring that everyone is a freemason from a sampling of regulars at the masonic lodge. Similarly, if some experiences don't affect downstream cognition, you won't be able to reflect on or recall them. There's a methodological paradox in inferring that all experiences are knowable or accessible from a sample of experiences guaranteed to be among the known and accessed ones.

Methodological paradox doesn't infect the other eight possibly essential features quite as inevitably, but sampling just from the masonic lodge remains a major risk. For example, even if it seems to you now that every experience you can introspect or remember constitutes a felt unity with every other experience had by you at the same moment, that could be an artifact of what you introspect and remember. Introspection might create unity where none was before. Disunified experiences, if they exist, might be quickly forgotten – never admitted to the masonic lodge. Similarly perhaps for indeterminate experiences, inflexible experiences, or atemporal experiences.

[51] See also Titchener 1901-1905; Hurlburt and Schwitzgebel 2007; Siewert 2007; Schwitzgebel 2011.

In principle, nonessentiality is easier to establish. A single counterexample suffices. One disunified, atemporal, or indeterminate experience would establish the nonessentiality of unity, specious presence, or determinacy. However, Problem One still applies. Accurately introspecting structural features of this sort is a surprisingly difficult enterprise.

*3. Introspection, Problem Three: The Narrow Evidence Base.*

The gravest problem lies in generalization beyond the human case. Waive worries about unreliability and sampling bias. Assume that you have correctly discerned through introspection and memory that, say, six of the ten proposed features belong to all of your experiences. Go ahead and generalize to all ordinary adult humans. It still doesn't follow that these features are universal among all possible experiencers. Maybe lizards or garden snails have experiences that lack luminosity, subjectivity, or unity. Since you can't crawl inside their heads, you can't know by introspection or experiential memory. (In saying this, am I assuming privacy? Yes, relative to you and lizards, but not as a universal principle.)

Even if we could somehow reasonably generalize from universality in humans to universality among animals, it wouldn't follow that those same features are universal among AI cases. Maybe AI systems can be more disunified than any conscious animal. Maybe, in defiance of privacy, AI systems can be built to directly introspect each other's experiences, without thereby collapsing into a single unified subject. Maybe AI systems needn't have the impression of the wonderful irreducibility of consciousness. Maybe some of their experiences could arise from reflexes with no possible downstream cognitive consequences.

Simple generalization from the human case can't warrant claims of universality across all possible conscious entities. The reason is fundamentally another version of sampling bias: Just as a biased sample of *experiences* can't warrant claims about all experiences, so also a biased sample of *experiencers* can't warrant claims about all experiencers. To defend the view that all conscious systems *must* have one or more of luminosity, subjectivity, unity, access, intentionality, etc., will require sturdier grounds than generalization from human cases.

*4. Conceptual Arguments, Problem One: The Shared Concept.*

Conceptual arguments don't rely on generalization from cases, so they are immune to concerns about sampling bias or a narrow evidence base. A conceptual argument for essentiality would attempt to establish that the feature is entailed by the very concept of consciousness. At the beginning of Chapter Three, I suggested that mentality and temporality are conceptually entailed essential features of consciousness. Consider also some other conceptual entailments: *Rectangle* entails *having four sides*. *Bachelor* entails *unmarried*. All *blue* things are also *colored*. All *trees* are *biological organisms*.

Chapter Two proposed that there's a standard, shared concept of (phenomenal) consciousness that we naturally grasp by considering examples and evocative phrases. If this shared concept exists, a challenge arises for anyone who holds that any of the ten features is entailed by that shared concept: Why do many philosophers and psychologists deny these entailments? Why aren't luminosity, subjectivity, etc., as obviously entailed by consciousness as four-sidedness is by rectangularity and coloration is by blueness? The explanation can't be *introspective* failure. We cannot say: The luminosity and subjectivity of experience are easy to miss because they are always present and thus easily ignored, unnoticed like a continual background hum. The method at hand isn't introspection. It's conceptual analysis. I examine my concept of consciousness. I attempt to discern its components and implications. I cannot discover a conceptual entailment to any of the ten features.[52]

I might be failing to see a subtle or complicated entailment. The concept of rectangularity entails that the interior angles sum to 360 degrees in a Euclidean plane. Without a geometrical education, this particular entailment is easily missed. Might luminosity, subjectivity, etc., be nonobvious conceptual entailments?

I cannot rule that out, but I can offer an account of how one might easily make the opposite mistake – the mistake of overattributing conceptual entailments.

Consider rectangularity again. Having two pairs of parallel sides might seem to be an essential feature. But it is not. In non-Euclidean geometry, rectangles needn't have parallel sides. It's understandable how someone who contemplates only Euclidean cases might mistakenly treat parallelism as essential. They might even form a nearby concept – rectangle-in-a-Euclidean-plane – which does have parallelism as an essential feature. But that is not the shared standard concept of rectangularity, at least in formal geometry.

Similarly, then, someone might regard luminosity, subjectivity, unity, etc., as essential features of consciousness if they consider only luminous, subjective, or unified cases. They might fail to consider or imaginatively construct possible cases that lack these properties, especially if such cases are unfamiliar. But AI cases might be to human cases as non-Euclidean geometry is to Euclidean geometry.

[52] On a sufficiently vacuous notion of subjectivity – perhaps Hume's (1740/1978) bundle view or Strawson's (2008) "thin" subject – subjectivity might be an exception. However, see note 37 on the distinction between the experience of subjectivity and the existence of a subject.

Thinking too narrowly, an advocate of the essentiality of one of these ten features might form a concept adjacent to the concept of consciousness, such as consciousness-with-luminosity, consciousness-with-subjectivity, consciousness-with-unity, etc. However, none of these concepts is the same as the concept of consciousness: There is no redundancy between the first and second parts of the concept as there is in rectangularity-with-four-sides. Rectangularity and rectangularity-with-four-sides really are the same concept. In Chapter Two, I defined consciousness by example, inviting you to notice the obvious shared property that is present in visual experience, pain, deliberate thought and imagery, silently humming a tune, and vivid emotion, but absent from early visual processing, the unnoticed processes guiding subtle changes in facial expression, and inactive memories. Assuming that consciousness and consciousness-with-luminosity are different concepts, which is the obvious one, most naturally picked out by the examples? I submit that it is consciousness plain, rather than consciousness-with-luminosity. Similarly for consciousness-with-subjectivity, consciousness-with-unity, consciousness-with-access, etc. The latter concepts are less simple. They involve extra, nonobvious components. Similarly, consciousness and consciousness-within-a-billion-kilometers-of-the-Earth's-surface will fit all the cases, but no one would treat the second property as the obvious one. Either *consciousness-with-X* is simply redundant with *consciousness*, which it doesn't appear to be, or it is a more complicated and less obvious referent than consciousness plain.

The argument I've just offered is, I recognize, hardly conclusive. I present it only as an explanatory burden that a defender of essentiality must meet.

*5. Conceptual Arguments, Problem Two: Nonobviousness.*

Most conceptual arguments for essential features of consciousness treat the essentiality as obvious on reflection. In my judgment, the essentiality is never as obvious as in claims like *bachelors cannot be married* or *blue is a color*. I'll present two influential examples to give a flavor.

*5.1. Example 1: Higher Order Thought and Luminosity.*

In his canonical early formulation and defense of the Higher Order Thought theory of consciousness, David M. Rosenthal writes:

> Conscious states are simply mental states we are conscious of being in. And in general our being conscious of something is just a matter of our having a thought of some sort about it. Accordingly, it is natural to identify a mental state's being conscious with one's having a roughly contemporaneous thought that one is in that mental state (2005, p. 26).[53]

One might interpret this as a conceptual argument. The concept of a conscious mental state is just the concept of a state we are conscious of being in, which in turn is just a matter of having an (unmediated and properly caused, as Rosenthal later clarifies) thought about that mental state. A type of luminosity (recall from Chapter Three) is therefore essential to consciousness: Consciousness conceptually entails ("higher order") knowledge of or awareness of or representation of some aspect of one's own mind.[54]

[53] Similarly, Lycan 2001.

[54] Though see Rosenthal on the possibility of mistakes. It is clear that Rosenthal intends this account to apply not only to humans but also at least to non-human animals. The seeming implication is that only animals with fairly sophisticated cognitive abilities (the ability to think about their mental states) can be conscious: Gennaro 2012; Rolls 2019.

Higher Order theories of consciousness are among the leading scientific contenders (see Chapter Eight). But few readers of Rosenthal – and perhaps not Rosenthal himself – regard this conceptual argument as sufficient on its own to establish the truth of Higher Order theory. Higher Order theorists typically seek empirical support. If the purely conceptual argument were successful, empirical support would be as otiose as polling bachelors to confirm that all bachelors are unmarried.[55]

Here's one reason Rosenthal's argument won't work purely as a conceptual argument: The term "conscious" can be understood either epistemically or experientially. Saying that I am *conscious of* something can be a way of saying I know something about it. Alternatively, it can be a way of saying that I'm having an experience of some sort. These meanings are linked, and it's natural to slide between them, since at least in familiar adult human cases, experiencing something normally involves knowing something about it. However, it is not evident as a matter of conceptual necessity that the epistemic and experiential need to be linked in the manner Rosenthal suggests, always and for all possible entities. The superficial appearance of a simple conceptual argument collapses if the experiential and epistemic senses of "conscious" are distinguished. "[Experientially] conscious states are simply mental states we are [epistemically] conscious of being in" might be true, but it is not a self-evident tautology.

Consciousness *might* essentially be luminous, or involve representation of one's own mental states, but a simple conceptual, linguistic argument of this sort doesn't establish that.

### *5.2. Example 2: Intentionality and Brentano's Thesis.*

---

[55] Brown 2025 especially emphasizes that the truth or falsity of Higher Order views will be decided on empirical rather than conceptual grounds.

Nineteenth-century philosopher and psychologist Franz Brentano famously argued that all mental phenomena are intentional, that is, are directed toward or about something:

> Every mental phenomenon is characterized by what the Scholastics of the Middle Ages called the intentional (or mental) inexistence of an object, and we might call, though not wholly unambiguously, reference to a content, direction toward an object (which is not to be understood here as meaning a thing), or immanent objectivity. Every mental phenomenon includes something as object within itself, although they do not all do so in the same way. In presentation, something is presented, in judgement something is affirmed or denied, in love loved, in hate hated, in desire desired, and so on.
>
> This intentional inexistence is characteristic exclusively of mental phenomena. No physical phenomenon exhibits anything like it. We can, therefore, define mental phenomena by saying that they are those phenomena which contain an object intentionally within themselves.[56]

Brentano's argument is conceptual. All judgments are judgments *about* something. Plausibly this is entailed by the very concept of a judgment. Loving likewise appears conceptually to entail an object – someone or something loved. If similar entailments hold for every possible mental state, then it is a conceptual truth that all mental states are intentional. Michael Tye's later argument that all mental states have "representational" content has a similar structure.[57]

The success of such arguments depends on the nonexistence of counterexamples, and since the beginning counterexamples have been proposed. Brentano discusses feelings such as pleasure. Tye discusses diffuse moods. Not only, the objector argues, can I be happy *about* something but I can also be happy *in general*, with no particular object. Brentano suggests that feelings without objects are about themselves.[58] Tye suggests that they represent bodily states.[59]

[56] Brentano 1874/1973, p. 88-89.

[57] Tye 1995.

Brentano or Tye might or might not be right about feelings and moods, but a disadvantage of approaching the conceptual question by enumerative example is that it's unclear on what grounds Brentano and Tye can generalize beyond the human case to all possible experiences by all possible experiencers. This variety of conceptual argument thus risks the same methodological shortcoming that troubles purely introspective arguments. Even granting that all *human* experience is intentional, that is a narrow base for generalizing to all possible experiencers, including novel AI constructs designed very differently from us. Brentano and Tye might be correct, but an enumerative conceptual argument alone cannot deliver the conclusion.

Some conceptual claims are obvious. In holding that bachelors are necessarily unmarried, we stand on solid ground. No similarly obvious conceptual argument supports the essentiality of any of the ten possibly essential properties of consciousness.

*6. Conceptual Arguments, Problem Three: Imaginative Limitation.*

One way to test for conceptual necessity is to seek imaginative counterexamples. If a thorough search for counterexamples yields no fruit, that's tentative evidence in favor of necessity. Of course, thoroughness is crucial. The advocate of the conceptual entailment from rectangularity to parallel sides failed to be thorough by neglecting non-Euclidean cases.

---

[58] Brentano 1874/1973, p. 90.

[59] Tye 1995, p. 124-129.

Our imaginations are limited. Moreover, we sometimes employ standards of successful imagination that illegitimately foreclose genuine possibilities. Consider another mathematical example: imaginary numbers. Ask a ten-year-old if they can imagine a number that doesn't fall on the real number line from negative to positive infinity. No, the student might say. Ah, but here comes *i,* the square root of negative 1. Suddenly, there's a whole world of imaginary and complex numbers that the middle-schooler had not thought to imagine. At first, before adjusting to the concept, the middle schooler might deny *i*'s imaginability. If the standard of successfully imagining a number N is imagining counting N beans or picturing N sheep, even negative numbers will seem unimaginable.

I advise considerable skepticism about claims of the unimaginability or inconceivability of conscious experiences lacking the ten possibly essential features. For example, you might struggle to conceive of a conscious experience without a sense of a subjective perspective (contra subjectivity), or an intermediate state between conscious and nonconscious (contra determinacy), or a partly disunified state where experience A is felt to co-occur with experience B and experience B with experience C but not A with C (contra unity). However, these difficulties might reflect constraints on what you're inclined to regard as a successful act of imagination, like our middle-schooler feeling that they need to picture some beans. Paradox ensues if the only permissible way to imagine a subjectless, indeterminate, disunified experience is as a vividly present experience in the unified field of an entity who feels like a subject.[60]

To escape imaginative ruts, consider some architectural facts about possibly conscious entities. For example, if consciousness depends on big, messy brains, it's unlikely always to switch on and off instantaneously, suggesting borderline cases in development, evolution, sleep, and trauma, contra determinacy. If we could design, build, breed, or discover a conscious entity with only partly unified cognition (maybe the octopus is an actual case), then consciousness too might be only partly unified.[61] If AI or organic systems could be conscious while directly accessing each other's interior structures, privacy might fail. I present these considerations not as full arguments but rather to loosen ungrounded presuppositions masquerading as conceptual necessities.

[60] For a detailed discussion, see Schwitzgebel 2023b; Schwitzgebel and Nelson 2025.

Some or all of these ten features of consciousness might indeed be essential. My argument so far is only that introspection and conceptual analysis alone cannot establish this. At best, they are weakly suggestive. We'll need, probably, to do some empirical science. But it's also hard to see how to resolve these issues empirically, as I'll discuss later.

Without clarity about the essential features of consciousness, we lose a crucial foothold for evaluating AI systems with architectures very different from our own. We know that if an AI is conscious, there must be "something it's like" to be them, but we won't know whether they need to represent their own processes, have unified cognition, have information widely accessible across the whole system, have a sense of self or of time, and so on – much less what specific *kinds* of self-representation, information-sharing, sense of self, etc., they would need to have.

---

[61] One real life case might be the Hogan twins, conjoined twins connected at the head, with overlapping brains and the capacity to report at least some of what is going on in each other's minds. See the 2017 CBC documentary *Inseparable: Ten Years Joined at the Head*. For other ways in which assumptions about AI unity might plausibly be violated or difficult to assess, see Birch 2025; Register 2025.

## Chapter Five: Materialism and Functionalism

Having covered some conceptual ground, let's step back for a broader metaphysical view. Are there compelling general metaphysical reasons (reasons, that is, pertaining to the fundamental structure of reality) to deny consciousness to AI systems, given what we *do* know about consciousness? In this chapter, I'll suggest probably not, unless one adopts a metaphysical view well outside of the scientific mainstream, and in most cases not even then.

### *1. Materialism Is Broadly Friendly to the Possibility of AI Consciousness.*

According to *materialism* (or *physicalism*), every concrete entity is composed of, reducible to, or most fundamentally, material or physical stuff – where "material or physical stuff" means things like elements of the periodic table and the various particles, waves, or fields that interact with or combine to form them. In particular, no immaterial soul exists and no mental properties exist distinct from that material or physical stuff. Your mind is somehow just a complex swirling of electrons, protons, and the like.[62]

Broadly speaking, materialism is friendly to the possibility of AI consciousness. At the deepest level, people and artificial machines don't differ, as they would if you had a soul while a machine did not. Although it might seem strange – maybe even inconceivable from our limited perspective[63] – that genuine consciousness could arise from electrical signals shooting across silicon wafers, consciousness does in fact arise from electrochemical signals shooting through neurons. If the latter is possible, the former might be too.

[62] On the challenges of defining materialism or physicalism, see Montero 1999; Stoljar 2010.

[63] Chalmers 1996; McGinn 2000.

Materialist arguments can be made against AI consciousness, at least on a moderately narrow definition of "AI". But since you and a robot are made fundamentally of the same basic stuff, those arguments must hinge on the specific material configurations involved, not on metaphysical dissimilarity at the most fundamental level.

*2. Alternatives to Materialism Don't Rule Out AI Consciousness.*

Materialism has been the dominant view in the natural sciences and mainstream Anglophone philosophy since at least the 1970s. I will assume it in the remainder of this book. However, alternatives remain worth considering. It's worth pausing to note that none of the main alternatives, in general form, disallows AI consciousness.

*Substance dualism* holds that mind is one type of thing, matter another. As Alan Turing noted (we'll return to him in Chapter Six), nothing in principle seems to prevent either God (by miracle) or a natural developmental process from instilling a soul in a computational machine.[64]

*Property dualism* holds that mental *properties* (such as the property of experiencing pain) are one thing, material properties (such as the property of undergoing such-and-such neural activity) another. Again, nothing in principle seems to prevent mental properties from arising in AI systems, and the most prominent advocate of property dualism, David Chalmers, defends the possibility of AI consciousness.[65]

According to *panpsychism,* consciousness is all-pervasive at the fundamental level of reality: Elementary particles and/or the cosmos as a whole is conscious. This might suggest liberality about AI consciousness. However, many panpsychists deny that middle-size aggregates such as rocks have conscious experiences distinct from the individual experiences of the particles composing them. Thus, panpsychism permits the same variable opinions about intermediate-sized objects as do other views.[66]

[64] Robinson 2003/2023; Turing 1950, p. 443.

[65] Chalmers 1996.

According to *metaphysical idealism,* there is no material world at all. Everything is fundamentally mental – just souls and their experiences. Behind our sensory experiences of rocks, trees, and tables stand no independently existing, material rocks, trees, or tables. AI systems might then also depend on patterns of experience in our minds. Yet minds must arise somehow – whether through natural law or divine action. Nothing in principle seems to preclude souls who experience artificial rather than biological embodiment.

According to *transcendental idealism,* fundamental reality is unknowable and (contra materialism, with some similarity to metaphysical idealism) the spatial features of reality depend on our minds. This epistemically modest view is entirely consistent with AI consciousness. Indeed, I've argued elsewhere that on one (simulationist) version of transcendental idealism, fundamental reality is a conscious computer.[67]

This list is not exhaustive, but the point should be clear: Rejecting materialism needn't entail rejecting AI consciousness.

*3. AI and the Spirit of Functionalism.*

What makes pain *pain?* Specifically, since we're now assuming materialism and interested in consciousness, what makes a particular material configuration a painful experience rather than hunger or no experience at all? Here you are, $10^{28}$ atoms spread through a wet, lumpy tenth of a cubic meter. What bestows the magic?

[66] Goff 2017; Roelofs 2019.

[67] Schwitzgebel 2017, 2024, ch. 5; Schlicht 2025.

The two most obvious and historically important materialist answers are: something about your material configuration or something about the causal patterns in which you participate. On the material configuration view, the reason you experience pain is that certain neurons in certain regions of your brain (or brain-plus-body or brain-plus-body-plus-environment) are active in a certain way.[68] On the causal patterns view – also known as *functionalism* – you experience pain because you are in a state that plays a certain causal or functional role in your cognitive economy (or the cognitive economy of your species). For example, you are in a state apt to have been caused by tissue stress and that is apt to cause in turn (depending on other conditions) avoidance, protection, anger, regret, and calls to the doctor.[69]

If the material configuration view is correct in its simplest form, then no neurons means no pain. Conscious states require biological neurons – or at least something sufficiently similar.[70] If artificial "neurons" don't count, then near-term AI is unlikely unless biological AI advances swiftly. We will discuss *biologicist* views in Chapter Ten.

In contrast, if functionalism is correct, then any computational system that implements the right causal/functional relationships will be conscious. Mental states can be understood entirely in terms of causal relationships to inputs and outputs and (crucially) among each other, requiring a certain internal causal architecture, but without essential reference to the particular materials out of which the system is composed. We will discuss some specific functionalist theories in Chapters Eight and Nine, but here I only want to highlight that functionalism is generally friendly in principle to the possibility of AI consciousness.

[68] Smart 2000/2022.

[69] Levin 2004/2023.

[70] Similar in what respect is a thorny issue, since structural similarity can be assessed in various respects, along various dimensions, at different scales, and with different coarseness of grain. "No neurons means no pain" is a simplification. See Bechtel and Mundale 1999.

The most common defense of functionalism is the *multiple realizability argument.* Humans feel pain but so also, plausibly, do octopuses, despite very different nervous systems.[71] If alien life exists elsewhere in this vast cosmos, as most astronomers think likely, then some aliens might also feel pain, despite radically different architectures. If so, pain can't depend too sensitively on specific details of material configuration.

A thought experiment: Tomorrow, flying saucers arrive. Friendly aliens disembark, speaking English and eager to converse with us about philosophy, psychology, space-faring technology, and the history of dance. When injured, they cry out, protest, protect the affected area, flap their antennae in distress (which they say is their equivalent of tears), seek medical help, avoid such situations in the future, and swear revenge. It seems natural to suppose that these aliens feel pain. (Chapter Ten will present an argument for this claim; for now, treat it as intuitive.)

But maybe inside they have nothing like human neurons. Maybe their cognition runs through hydraulics, internal capillaries of reflected light, or chemical channels. What matters, the functionalist says, is not *what they're made of* but rather *how they function.* Do they receive input from the environment and respond to it flexibly in light of past events? Do they preserve themselves over time, suffering short-term losses to avoid larger long-term risks? Do they communicate detailed information with each other? Do they monitor their internal processes, report them to others, and integrate inputs from a variety of sources over time to generate intelligent action? If they have enough of the right sort of these functional processes, then they are conscious, regardless of what they happen to be made of.

[71] Bickle 1998/2020.

Functionalist philosophers and psychologists approach AI with the same liberality, focusing on whether systems implement the right functional processes, regardless of their material composition. The question is only what specific functions are sufficient for consciousness and how close our current systems are to implementing those functions.

Unsurprisingly, the answer depends in part on the ten contested essential features of consciousness described in Chapter Three.

*4. Computational Functionalism.*

According to *computational functionalism,* mentality is computation. The functional processes constitutive of the mind are *computational* processes. In principle, this position is even more hospitable to AI consciousness than functionalism generally. Whatever computational processes suffice for consciousness in us, if we can reproduce them in AI, then that system will be conscious.

But what is computation? On a very liberal view, any process can be described computationally, in terms of abstract if-then rules. Cars zipper-merging on a freeway can be described computationally as converting 0, 0, 0… in the left lane and 1, 1, 1… in the right lane into 0, 1, 0, 1, 0, 1… in the merged lane. An acorn dropping from a tree can be described as a process of subtracting one from the sum of acorns on the tree and adding one to the sum on the ground.[72] It's then trivially true that whatever processes generate consciousness in us can be described computationally.

[72] For defenses of pancomputationalism: Putnam 1967; Searle 1992; Chalmers 1996.

Critics object that description is not creation. A computational model of a hurricane gets no one wet; a computational model of an oven cooks no turkey. Similarly, a computational model of a mind, even if executed in complete detail on a computer, might not generate consciousness.[73] Proponents of computational functionalism can reply that the mind is different, computation being its essence. Alternatively – retreating from the strongest version of computational functionalism – defenders of AI consciousness can note that AI systems have sensors, effectors, and real physical implementations. If they emit microwaves, they *can* cook turkeys. Their reality isn't exhausted by their computational description. Even if computation alone isn't enough, maybe the right computations plus the right sensors, effectors, and implementations would suffice for consciousness.

A narrower definition of computation, advanced by Gualtiero Piccinini, restricts computation to systems with the function of manipulating "medium-independent vehicles" according to rules – where medium-independent vehicles are physical variables defined solely in terms of their degrees of freedom (e.g., 0 vs. 1) rather than their specific physical composition.[74] Maybe human brains perform computation in that sense; maybe not.[75] Without entering into the details, we can again note that AI systems do more than just compute. They can output readable text and manipulate real physical objects via effectors. So one needn't hold that the right type of computation is sufficient by itself for consciousness to hold that AI systems might be conscious.

Conversely, even if computational functionalism is true, that's no guarantee that it's possible to instantiate the relevant computations on any feasible AI system in the foreseeable future.

---

[73] Searle 1984.

[74] Piccinini 2015.

[75] Piccinini and Bahar 2013; Rescorla 2015/2025.

## Chapter Six: The Turing Test and the Chinese Room

This chapter evaluates two influential arguments about near-term AI consciousness: one in favor, based on the "Turing test", and one against, based on John Searle's "Chinese room" and Emily Bender's related "underground octopus". One advantage of these arguments is that they don't require commitment to the essentiality or inessentiality of any of the ten features discussed in Chapter Three. One disadvantage is that they don't work.

### *1. Against the Turing Test as an Indicator of Consciousness.*

It's tempting to think that sufficiently sophisticated linguistic behavior warrants attributing consciousness. Imagine the aliens from Chapter Five emitting sounds or text that we naturally interpret as English sentences, with the apparent acuity of an educated human. In the spirit of functionalist liberalism about architectural details, one might regard this as sufficient to establish consciousness, even knowing nothing about their bodies, internal structures, or non-linguistic behavior.

Alan Turing's 1950 "imitation game" – better known as the Turing test – treats linguistic indistinguishability from a human as sufficient grounds to attribute "thought".[76] If a machine's verbal behavior is sufficiently humanlike, we should allow that it thinks. This idea has been adapted (contra Turing, as I'll discuss below) as a test of consciousness.[77]

In Turing's original setup, a human and a machine, through a text-only interface, each try to convince a human judge that they are human. The judge is free to ask whatever questions they like, attempting to prompt a telltale nonhuman response from the machine. The machine passes if the judge can't reliably distinguish it from the human. More broadly, we might say that a machine "passes" if its verbal outputs strike users as sufficiently humanlike to make discrimination difficult.

[76] Turing 1950.

[77] Harnad 2003; Schneider 2019.

Indistinguishability comes in degrees. Turing tests can have relatively high or low bars. A low-bar test might involve:

- *ordinary users* as judges, with no special expertise;
- *brief interactions,* such as five minutes;
- *a relaxed standard of distinguishability,* for example, the machine passes if 30% of judges guess wrong.

A high-bar test might require:

- *expert judges* trained to distinguish machines from humans;
- *extended interactions,* such as an hour or more;
- *a stringent standard of distinguishability,* for example, the machine fails if 51% of judges correctly identify it.

The best current language models already pass a low-bar test.[78] But language models might not pass high-bar tests for a long time, if ever. So let's avoid talk about whether machines pass "the" Turing test. There is no one Turing test.

A better question is: *What type and degree of Turing indistinguishability, if any, would establish that a machine is conscious?* Indistinguishability to experts or non-experts? Over five minutes or five hours? With what level of reliability? We might also consider topic-relative or tool-relative indistinguishability. A machine might be Turing indistinguishable (to some judges, for some duration, to some standard) when discussing sports or fashion but not when discussing consciousness.[79] A machine might fool unaided judges but fail when judges employ detection tools.

[78] Jones and Bergen 2025.

Turing himself proposed a relatively low bar:

> I believe that in about fifty years' time it will be possible, to programme computers... to make them play the imitation game so well that an *average interrogator* will not have more than *70 per cent chance* of making the right identification after *five minutes* of questioning… [and] one will be able to speak of machines thinking without expecting to be contradicted.[80]

I have italicized Turing's implied standards of judge expertise, indistinguishability, and duration.

However, regarding consciousness, Turing writes:

> I do not wish to give the impression that I think there is no mystery about consciousness. There is, for instance, something of a paradox connected with any attempt to localise it. But I do not think these mysteries necessarily need to be solved before we can answer the question with which we are concerned in this paper.[81]

Turing thus set aside the question of consciousness. This is, I think, wise. Whether it's reasonable to describe a machine as "thinking", "wanting", "knowing", or "preferring" one thing or another is to some extent a matter of practical convenience. Consider a language model integrated into a functional robot that tracks its environment and has specific goals. As a practical matter, it will be difficult to avoid saying that the robot "thinks" that the pills are in Drawer A and that it "prefers" the slow, safe route over the quick, risky route, especially if it verbally affirms these opinions and desires. It will be irresistibly convenient to speak that way even if, strictly speaking, such entities lack whatever it takes to really have beliefs, desires, and thoughts (a tricky issue we can't explore here, which might be partly dissociable from the question of whether they are conscious).[82]

[79] Turner and Schneider's test turns specifically on the machine's ability to discuss consciousness: Schneider 2019, though for concerns see Udell and Schwitzgebel 2021.

[80] Turing 1950, p. 442.

[81] Turing 1950, p. 447.

The main question at hand does not concern practical matters of terminological convenience. It concerns a matter of fact independent of how we happen to speak: whether near-future AI systems might *actually have conscious experiences*.

For consciousness, we should probably abandon hope of a Turing-test standard.

Note, first, that it's unrealistic to expect any near-future machine to pass the very highest bar Turing test. No machine is likely to reliably fool experts who specialize in catching them out, who are armed with unlimited time and tools, and who need to exceed 50% accuracy by only the slimmest margin. As long as machines and humans differ in underlying architecture, they will differ in their patterns of response in some conditions, which experts can be trained or equipped to detect.[83] To insist on an impossibly high standard is to guarantee in advance that no machine could prove itself conscious, contrary to the spirit of the test. Imagine applying such a ridiculously unfair test to a visiting space alien.

---

[82] This seems especially likely on interpretativist, fictionalist, and antirealist views about belief: Dennett 1987; Mölder 2010; Toon 2023; Schwitzgebel forthcoming; see also Cappelen and Dever 2025. Turing might be similarly pragmatic and antirealist about thinking: He mostly uses the term "think" in scare quotes. At one point he writes, "The original question, 'Can machines think?' I believe to be too meaningless to deserve discussion" (1950, p. 442).

[83] Barring fantastical superemulators; barring giving the machines a chance to train against those experts in a mimic vs. dupe arms race; barring the machines being informed of the experts' techniques.

Too low a bar is equally unhelpful. As noted, machines can already pass some low-bar tests, despite lacking the capacities and architectures that most experts think are necessary for consciousness. To assume without substantial further argument that a low-bar Turing test establishes consciousness contradicts almost every scientific theory and the majority of experts on the topic.

Could we choose just the right mid-level bar – high enough to rule out superficial mimicry, low enough not to be ridiculously unfair? I see no reason to think that there must be some "right" level of Turing indistinguishability that reliably reveals consciousness. The past seven years of language-model achievements suggest that with clever engineering and ample computational power, superficial fakery might bring a nonconscious machine past any reasonably fair Turing standard. (For more on AI mimicry, see Chapter Seven.)

Turing indistinguishability is an interesting concept with a variety of potential implications – for example, in customer service, propaganda production and detection, and AI companions. But for assessing consciousness, we'll want to look beyond outward linguistic behavior.

*2. The Chinese Room and the Underground Octopus.*

In 1980, John Searle proposed a thought experiment: He is locked in a room and receives Chinese characters through a slot. Unfamiliar with Chinese, he consults a massive rulebook, following detailed instructions for arranging and rearranging those characters alongside a store of others, eventually passing new characters back through the slot. Outside the room, people interpret the inputted characters as questions in Chinese and the outputted characters as responses. With a sufficiently large and well-written rulebook, and ignoring time constraints, it might appear from outside as if Searle is conversing in Chinese.[84]

Searle argues that if AI programs consist of if-then rules (as in Turing's standard model of digital computation[85]), then in principle he could instantiate any AI program in this manner. But neither he nor the larger system of man-plus-rulebook-plus-room understands Chinese. Therefore, Searle concludes, even if a computer program could produce outputs indistinguishable from those of a Chinese speaker, this is insufficient for genuine understanding. Searle's original 1980 article doesn't address consciousness, but his subsequent work makes clear that he intends the argument to work for consciousness also.[86]

The Chinese room argument has generated extensive debate, much of it critical.[87] Some of the skepticism is justified – and the reader will notice that I have not rested my argument against the Turing test on Searle's criticism. In my assessment, the crucial weakness is the argument's reliance on the intuition – assertion? assumption? – that neither Searle nor any larger system of which he is a part knows Chinese.[88] It *might* be the case that nothing in the system would know Chinese, but Searle's argument is insufficient to establish that conclusion.

In imagining the thought experiment, you might picture Searle working slowly through a 2000-page tome, outputting sets of characters every several minutes. And it does seem plausible if *that* were the procedure, nobody knows Chinese. But to actually pass a medium-bar Turing test, the setup would need to be vastly more powerful. Our best large language models, the ones that pass low-bar Turing tests, execute *hundreds of trillions of instructions* in dealing with complex input-output pairs. To match that, Searle would need tens of thousands of human lifetimes' worth of error-free execution. Alternatively, we might imagine a single giant lookup table with one page for every possible five-minute input sequence and its corresponding output. If we assume 3000 possible Chinese characters at one character input per second for five minutes, the rulebook would require approximately $10^{1000}$ pages – many orders of magnitude more pages than there are atoms in the observable universe. *Maybe* no Chinese would be understood in the process; but that requires an argument. Human intuitions adapted for familiar cases might be as ill-suited to procedures of that magnitude as intuitions based on tossing rocks are ill-suited to evaluating the behavior of photons crossing the event horizons of black holes.

[84] Searle 1980; relatedly, Block 1981.

[85] Turing 1936.

[86] Searle 1992.

[87] See the replies in Searle 1980; Cole 2004/2024.

[88] See the "systems reply" in Searle 1980; Cole 2004/2024.

This isn't to say that Searle or the system to which he contributes *would* understand – just that we shouldn't confidently assume that our impressions based on familiar cases should extend to the Chinese room case conceived in its proper magnitude. In fact, as I will argue in the next chapter, if the Chinese room was designed specifically to mimic the superficial features of human linguistic output, there's good reason to be skeptical about its outward signs of consciousness. That argument – the Mimicry Argument – is grounded in the epistemic principle of inference to the best explanation rather than in an appeal to intuitive absurdity.

Emily Bender and colleagues develop a similar example in a pair of influential papers from 2020 and 2021.[89] Large language models, they say, are "stochastic parrots" that imitate human speech by detecting statistical relationships among linguistic items, reproducing familiar patterns without understanding. The most successful language models in 2020 – pure transformer models like GPT-3 – did indeed work like complex parrots: They tracked and recreated co-occurrence relationships among words or word-parts. Simplifying: If "peanut butter and" is usually followed by "jelly" in the huge training corpus of human texts, the model predicts and outputs "jelly" as the next word. Recycling that output as a new input, if "peanut butter and jelly" is usually followed by "sandwich", the model outputs "sandwich". And on it goes. Unlike autocomplete on ordinary phones at the time, these statistical relationships can bridge across intervening phrases: If "peanut butter and jelly sandwich" has been preceded by "I love a good", the model will predict and output a different next phrase than if it has been preceded by "Please don't feed me another". "Attention" mechanisms give words different weights in connection with other words, again patterned on human usage.

[89] Bender and Koller 2020; Bender et al. 2021.

More recent language models aren't quite so simply imitative. For example, in post-training they will receive feedback that makes certain outputs more likely and others less likely for reasons like safety and helpfulness – reasons, that is, other than matching patterns in the training corpus. But extensive training to match human word co-occurrence patterns remains at the core of the models' functionality.

Bender and colleagues invite us to imagine an underground octopus eavesdropping on a conversation conducted via cable between two people stranded on remote islands. Once the octopus has observed enough of their interaction, it can sever one end of the cable, substituting its own replies for those of the disconnected partner. It might fool the other island dweller for a while, passing a low-bar Turing test. But never having seen an island or a human, the octopus will not really understand what a coconut or a palm tree or a human hand is (this is sometimes called the "symbol grounding problem"[90]). Bender and colleagues suggest that the octopus's ignorance will be revealed when asked for specific help with a novel physical task, such as building a coconut catapult. Without understanding the meanings of the words and their relationships to everyday physics, it will be limited to responses like "great idea!" or suggestions unconstrained by physical plausibility.

[90] Harnad 1990.

Bender and colleagues might or might not be right about the octopus's limitations. Subsequent language models have done surprisingly well – surprising from the perspective of 2020, at least – on even seemingly novel tasks one might have thought would require understanding meaning and not just statistical relationships among lexical items. The models are still far from perfect, and it's very much up for debate whether their patterns of failure reveal a fundamental lack of understanding or only specific deficiencies. And regardless of the answer to that particular question, near-future AI systems needn't be "underground" like the octopus: They can be robotically embodied in natural environments, potentially sidestepping Bender's main argument.

Similarly, Searle explicitly restricted his argument to Turing-style digital computers, not to AI systems of very different architectures that might soon emerge (recall Chapter Two).[91] Even if his or Bender's arguments reveal the nonconsciousness of the best known current AI systems, they do not generalize to near-future AI in general.

Regardless, Bender's octopus, like Searle's Chinese room, lays its finger (arm tip?) on an important worry. If a system is designed specifically to mimic patterns in human speech, the best explanation of its apparent fluency might be that it is an excellent mimic, rather than that it possesses the structures necessary for genuine understanding or consciousness. Rightly, we mistrust mimics. Copying the surface does not entail copying the depths. Next, let's consider the Mimicry Argument in more detail.

[91] Searle 1980, p. 422.

## Chapter Seven: The Mimicry Argument Against AI Consciousness

This chapter presents what might be the best argument for skepticism about the consciousness of AI systems that are behaviorally very similar to us. The argument is inspired in part by Searle's and Bender's thought experiments, and it generalizes from passing remarks by many skeptics who hold that AI systems merely mimic, imitate, or simulate consciousness. However, it supports only the weak conclusion that superficial behavioral evidence doesn't justify positively attributing consciousness to "consciousness mimics". The Mimicry Argument does not establish the stronger conclusion that AI systems are demonstrably nonconscious.

### *1. Mimicry in General.*

In mimicry, one entity (the mimic) possesses a superficial or readily observable feature that resembles that of another entity (the model) because of the impact of that resemblance on an observer (the receiver), who treats the readily observable feature of the model as indicating some further feature. See Figure 1. For example, viceroy butterflies mimic monarch butterflies' wing coloration patterns to mislead predator species who avoid monarchs due to their toxicity.[92] An octopus can adopt the color and texture of its environment to seem to predators like an unremarkable (and inedible) continuation of that environment. Gopher snakes vibrate their tails in dry brush, mimicking a rattlesnake's rattle to deter threats.

[92] For some complications, see Prudic et al. 2019.

Figure 1: The mimic's possession of readily observable feature S2 is explained by its resemblance to feature S1 in the model because of how a receiver, who treats S1 as indicating further feature F, responds to the resemblance of S1 and S2. S1 reliably indicates F in the model but S2 need not reliably indicate F in the mimic.[93]

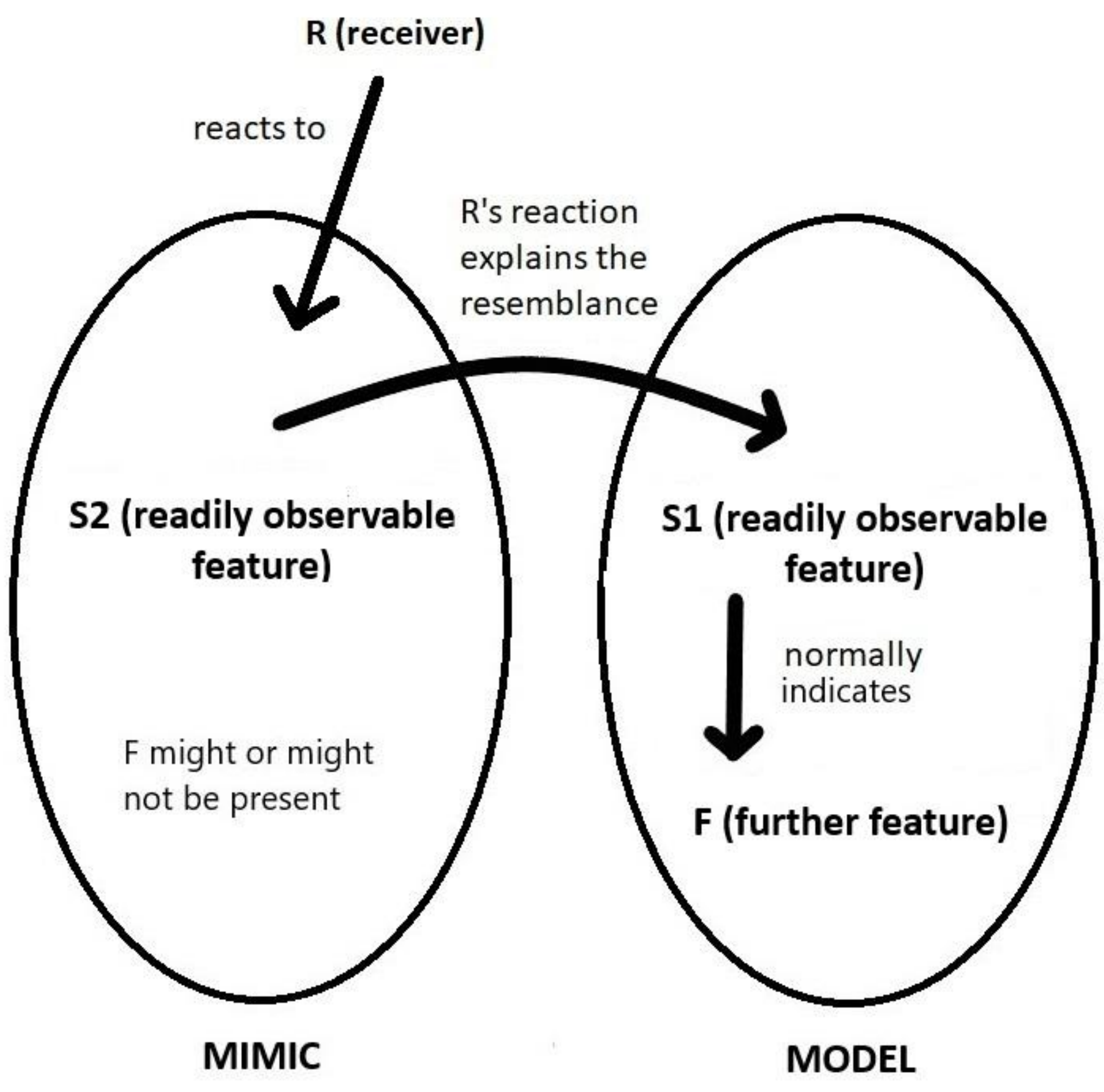

[93] Image source: Schwitzgebel & Pober forthcoming, from which the ideas of this section are drawn.

Not all mimicry is deceptive. Parrots mimic each other's calls to signal group membership, and they can do so either deceptively or non-deceptively. A street mime might mimic a depressed person's gait to amuse bystanders, who of course don't think the mime is depressed. Turning to a technological example, a simple doll might say "hello" when powered on, mimicking a human greeting.

Mimicry is more than simple imitation. Mimicry requires an targeted receiver, and that receiver must normally treat the readily observable feature, when it occurs in the model entity, as indicating some further feature: The parrot's call normally indicates group membership; the gait normally indicates a depressed attitude; the sound "hello", when spoken by the model entities (humans), normally indicates an intention to greet. This complex relationship between mimic, model, receiver, two readily observable features, and one further feature must be the reason that the mimic exhibits the readily observable feature in question. Ordinary imitation, in contrast, can have any of a variety of goals. For example, you might imitate someone's successful stone-hopping to avoid wetting your feet in a stream, or you might imitate the bench press form of a personal trainer to improve your own form. Unless there's a targeted receiver who reacts to the imitation because of what the feature normally indicates in the model – and whose reaction is the point of or explanation of the imitation – mimicry strictly speaking has not occurred.

We can also contrast mimicry with childhood language learning. Suppose a child learns a novel word ("blicket") for a novel object (a blicket), repeating that word in imitation of an adult speaker. The best explanation of their utterance is as a direct signal of their own knowledge that the object is a blicket, not the complex mimicry relationship.

When you know that something has been designed or has evolved as a mimic, you cannot infer from the readily observed feature to the further feature in the way you ordinarily would in the model. At least you can't do so without further evidence. Once you know that the viceroy mimics the monarch, you cannot infer from its wing pattern to its toxicity. Maybe the viceroy is toxic, but that would need to be separately established. Similarly, knowing that the toy's "hello" mimics a human greeting, you cannot infer that the toy actually intends to greet you. Referring back to Figure 1, when confronted with the *model*, you can infer from readily observable feature S1 to further feature F, but when confronted with the *mimic*, you cannot infer from readily observable feature S2 to further feature F.

*2. The Chinese Room, the Underground Octopus, and the Mimicry Argument.*

Searle's Chinese room and Bender's underground octopus are mimics in this sense. Their readily observed features are their textual outputs, designed to resemble those of a human Chinese speaker or an island conversational partner. In humans, such outputs reliably indicate consciousness and linguistic understanding. But when those outputs arise from mimicry, we can't – at least not without further argument – infer consciousness or linguistic understanding. The inference from sophisticated text to underlying conscious experience is undercut.

More generally, the Mimicry Argument against AI consciousness works as follows. A *consciousness mimic* is an entity that mimics some superficial or readily observable features that, in some set of model entities, reliably indicate consciousness.[94] But because the mimic has been designed or selected specifically to display those superficial features, we the receivers cannot justifiably infer underlying consciousness – not in the same way we can when we see those same features in the model entity. This is obvious for the "hello" toy, less obvious but still true for entities specifically designed to pass the Turing test or otherwise mimic the surface features of human language. An important class of AI systems are consciousness mimics in this sense.

[94] A nuance: For consciousness mimicry so defined, feature F needn't itself be consciousness as long as S1 reliably indicates consciousness via indicating F in the model. For example, the F that the receivers care about might be semantic content.

Searle and Bender aim for a stronger conclusion, inviting us positively to conclude that the mimics *do not* have conscious linguistic understanding. I don't think we can know this from their arguments. But both thought experiments successfully describe consciousness mimics whose outputs we should reasonably mistrust. The case *for* consciousness is undercut. It does not follow that the case *against* consciousness is established.

Compare with classic examples from epistemology. Ordinarily, if you see a horse-shaped animal with black and white stripes in a zoo, you can infer that it's a zebra. But if you know that the zookeepers care only about displaying something with the superficial appearance of a zebra, good enough to delight naive visitors, you ought no longer be so sure. Maybe they've painted stripes on mules.[95] Ordinarily, if you see a barn-like structure in the countryside, you can infer the presence of a barn. But if you know that a Hollywood studio is filming nearby and cares only about creating the superficial appearance of a barn-studded landscape, you ought no longer be so sure. Some of the seeming-barns might be mere facades.[96]

Ordinarily, if you're having what seems to be a meaningful conversation, you can infer that your conversation partner is conscious and understands the meaning of your words. But if you know that the entity is designed to mimic human text outputs, you ought no longer be so sure.

---

[95] Dretske 1970. Note that an animal's zebra-like appearance might still be non-conclusive *evidence* that it is a zebra, rationally increasing one's credence. Analogously, one might treat S2 in the mimic as non-conclusive evidence of consciousness, for example, if one invests a non-zero credence in the hypothesis that the likeliest way to be an effective mimic is to actually be conscious.

[96] Goldman 1976.

*3. Consciousness Mimicry in AI.*

Classic pure transformers such as GPT-3, as described in Chapter Six, are consciousness mimics. They are trained to output text that closely resembles human text (the superficial feature), so that human receivers will interpret them as utterances with linguistic meaning (the further feature), and utterances with linguistic meaning normally entail consciousness in the model entities (humans). An important discovery of the late 2010s and early 2020s was that such mimics could fool ordinary users in brief interactions.[97] The Mimicry Argument straightforwardly applies. We cannot infer from the superficial text outputs to underlying consciousness. Any argument that such machines are conscious must appeal to further considerations. In the next chapter we'll begin to consider what such arguments might look like, but the large majority of experts on consciousness agree that classic pure transformers are not conscious.

Models programmed in a more traditional manner (through lines of code in a traditional programming language) can also be seen as consciousness mimics. The "hello" toy is a simple example. A slightly less simple example is a program designed to output sentences like "Please enter the dates you wish to travel" or "Thank you for flying with Gigantosaur Airlines!" Such text outputs are modeled on English speakers' linguistic behavior for the sake of a receiver who will attribute linguistic significance, but they don't reveal any understanding in the machine. The machine needn't have whatever underlying cognitive or architectural structures are necessary for genuine comprehension.

---

[97] Schwitzgebel, Strasser, and Schwitzgebel 2024; Fiedler and Döpke 2025; Jones and Bergen 2025.

Not all AI systems are consciousness mimics. AlphaGo, for example, was trained to play the game of Go by competing against itself billions of times, gradually strengthening connection weights leading to wins and weakening those leading to losses. By 2016, it was expert enough to defeat the world's best Go players.[98] Although some elements of its interface might involve linguistic mimicry, its basic functionality was not mimetic. It was trained to be good at Go, not just to have the superficial appearance of a Go player. Similarly, a calculator is not a mimic. It was designed to track arithmetic principles, not human behavior, though its outputs are shaped to be interpretable by human users. Of course, few people think AlphaGo or a calculator are conscious.

Recent Large Language Models such as ChatGPT and Claude build upon the mimicry structures of pure transformer models but also receive post-training. Reinforcement learning from human feedback "rewards" human-approved outputs, strengthening the associated weights. Models can also be reinforced for being "right" by external standards, and some can access tools like calculators. To the extent the machines move beyond pure mimicry, the Mimicry Argument applies less straightforwardly. For now, mimicry-based skepticism still seems warranted, since their core architecture remains close to that of pure transformers, and their humanlike outputs are still best explained by their pretraining on word co-occurrence in human texts.

In the longer term, we might imagine architectures more thoroughly trained on the rights and wrongs of the world itself – maybe like AlphaGo but with the larger world, or some significant portion of it, as its playground. Outputs would be shaped primarily by success in real-world complex tasks, perhaps including communicative tasks, rather than by resemblance to humans. The Mimicry Argument would then no longer apply. Skepticism about their consciousness, if warranted, would need a different basis.

[98] Silver et al. 2016; Mozur 2017.

Pulling together the threads of these past two chapters:

Perhaps with sufficient time and computational resources a machine could be designed to almost perfectly mimic human linguistic behavior, passing even a high-bar Turing test. Recent developments in AI have shown that, in practice, machines can fool ordinary users in brief conversations, with further improvements likely. However, if mimicry of human text patterns is the best explanation of the outputs, we cannot simply infer consciousness from their humanlike appearance, as we might with non-mimic entities like humans or (presumably) hypothetical space alien visitors. Knowing that the system is designed as a mimic undercuts the usual inference from superficial behavior to underlying conscious cause.[99]

[99] See also Birch 2024 on the “gaming problem”.

## Chapter Eight: Global Workspace Theories and Higher Order Theories

If superficial patterns of language-like behavior cannot by themselves establish that an entity is conscious, where else might we look? One answer is the functionalist's: Look to the functional architecture, that is, the patterns of causal relationships among inputs, outputs, and various internal states. In the broad spirit of functionalism, we shouldn't demand too specifically humanlike a design, with exactly the same fine-grained functional structures we see in ourselves. More plausibly, what matters are big-picture functional relationships – especially those linked to the ten possibly essential features of consciousness.

The leading candidate – though far from a consensus view – is some version of Global Workspace Theory.[100] Higher Order theories are also prominent and closely related.[101] This chapter examines both approaches.

### *1. Global Workspace Theories and Access.*

The core idea of Global Workspace Theory is simple. Sophisticated cognitive systems like the human mind employ specialized processes that operate to a substantial extent in isolation. We can call these *modules*, without committing to any strict interpretation of that term.[102] For example, when you hear speech in a familiar language, some cognitive process converts the incoming auditory stimulus into recognizable speech. When you type on a keyboard, motor functions convert your intention to type a word like "consciousness" into nerve signals that guide your fingers. When you try to recall ancient Chinese philosophers, some cognitive process pulls that information from memory without (amazingly) clogging your consciousness with irrelevant information about German philosophers, British prime ministers, rock bands, or dog breeds.

[100] E.g., Baars 1988; Dehaene 2014; Mashour et al. 2020.

[101] E.g., Rosenthal 2005; Gennaro 2012; Lau 2022; Brown 2025.

[102] Full Fodorian (1983) modularity is not required.

Of course, not all processes are isolated. Some information is widely shared, influencing or available to influence many other processes. Once I recall the name "Zhuangzi", the thought "Zhuangzi was an ancient Chinese philosopher" cascades downstream. I might say it aloud, type it out, use it as a premise in an inference, form a visual image of Zhuangzi, contemplate his main ideas, attempt to sear it into memory for an exam, or use it as a clue to decipher a handwritten note. To say that some information is in "the global workspace" just is to say that it is available to influence a wide range of cognitive processes. According to Global Workspace Theory, a representation, thought, or cognitive process is conscious if and only if it is in the global workspace – if it is "widely broadcast to other processors in the brain", allowing integration both in the moment and over time.[103] The content of this workspace is in constant flux. It includes most or all of what you're attending to and excludes most or all of what you're not attending to.

Recall the ten possibly essential features of consciousness from Chapter Three: luminosity, subjectivity, unity, access, intentionality, flexible integration, determinacy, wonderfulness, specious presence, and privacy. Global Workspace Theory treats *access* as the central essential feature.

Global Workspace theory can potentially explain other possibly essential features. *Luminosity* follows if processes or representations in the workspace are available for introspective processes of self-report. *Unity* might follow if there's only one workspace, so that everything in it is present together. *Determinacy* might follow if there's a bright line between being in the workspace and not being in it. *Flexible integration* might follow if the workspace functions to flexibly combine representations or processes from across the mind. *Privacy* follows if only you can have direct access to the contents of your workspace. *Specious presence* might follow if representations or processes generally occupy the workspace for some hundreds of milliseconds.

[103] Mashour et al. 2020, p. 776-777.

In ordinary adult humans, typical examples of conscious experience – your visual experience of this text, your emotional experience of fear in a dangerous situation, your silent inner speech, your conscious visual imagery, your felt pains – appear to have broad cognitive influences. It's not as though we commonly experience pain but find that we can't report it or act on its basis, or that we experience a visual image of a giraffe but can't engage in further thinking about the content of that image. Such general facts, plus the theory's potential to explain features such as luminosity, unity, determinacy, flexible integration, privacy, and specious presence, lend Global Workspace Theories substantial initial attractiveness.

I have treated Global Workspace Theory as if it were a single theory, but it encompasses a family of theories that differ in detail, including "broadcast" and "fame" theories – any theory that treats the broad accessibility of a representation, thought, or process as the central essential feature making it conscious.[104] Consider two contrasting views: Dehaene's Global Neuronal Workspace Theory and Daniel Dennett's "fame in the brain" view. Dehaene holds that entry into the workspace is all-or-nothing. Once a process "ignites" into the workspace, it does so completely. Every representation or process either stops short of entering consciousness or is broadcast to all available downstream processes. Dennett's fame view, in contrast, admits degrees. Representations or processes might be more or less famous, available to influence some downstream cognitive processes without being available to influence others. There is no *one* workspace, but a pandemonium of competing processes.

[104] E.g. Baars 1988; Dennett 1991, 2005; Tye 2000; Prinz 2012; Dehaene 2014; Mashour et al. 2020.

Whether Dennett's view is more plausible than Dehaene's turns on whether, or how commonly, representations or processes are *partly* famous. Some visual illusions, for example, seem to affect verbal report but not grip aperture: We *say* that X looks smaller than Y, but when we *reach* toward X and Y we open our fingers to the same extent, accurately reflecting that X and Y are the same size. The fingers sometimes know what the mouth does not (Aglioti et al. 1995; Smeets et al. 2020). We adjust our posture while walking and standing in response to many sources of information that are not fully reportable, suggesting wide integration but not full accessibility (Peterka 2018; Shanbhag 2023). Swift, skillful activity in sports, in handling tools, and in understanding jokes also appears to require integrating diverse sources of information, which might not be *fully* integrated or reportable (Horgan and Potrč 2010; Christensen et al. 2019; Vauclin et al. 2023). In response, the all-or-nothing "ignition" view can explain away such cases of seeming intermediacy or disunity as atypical (it needn't commit to 100% exceptionless ignition with no gray-area cases), by allowing some nonconscious communication among modules (which needn't be entirely informationally isolated), and/or by allowing for erroneous or incomplete introspective report (maybe some conscious experiences are too brief, complex, or subtle for people to confidently report experiencing them).

If Dennett is correct, luminosity, determinacy, unity, and flexible integration all potentially come under threat in a way they do not as obviously come under threat on Dehaene's view.[105] Dennettian concerns notwithstanding, all-or-nothing ignition into a single, unified workspace is currently the dominant version of Global Workspace Theory. The issue remains unsettled and has obvious implications for the types of architectures that might plausibly host AI consciousness.

*2. Consciousness Outside the Workspace; Nonconsciousness Within It?*

Global Workspace Theory is not the correct theory of consciousness unless *all* and *only* thoughts, representations, or processes in the Global Workspace are conscious. Otherwise, something else, or something additional, is necessary for consciousness.

It is not clear that even in ordinary adult humans a process must be in the Global Workspace to be conscious. Consider the case of peripheral experience. Some theorists maintain that people have rich sensory experiences outside of focal attention: a constant background experience of your feet in your shoes and objects in the visual periphery.[106] Others – including Global Workspace theorists – dispute this. Introspective reports vary, and resolving such issues is methodologically tricky.

One methodological problem: People who report constant peripheral experiences might mistakenly assume that such experiences are always present because they are always present *whenever they think to check,* and the very act of checking might generate those experiences.

[105] Despite developing a theory of consciousness, Dennett (2016) endorsed "illusionism", which rejects the reality of phenomenal consciousness (see especially Frankish 2016). I interpret the dispute between illusionists and nonillusionists as a verbal dispute about whether the specific philosophical concept of "phenomenal consciousness" requires immateriality, irreducibility, perfect introspectibility, or some other dubious property, or whether the term can be "innocently" used without invoking such dubious properties. See Schwitzgebel 2016, 2025; Kammerer 2025.

[106] Reviewed in Schwitzgebel 2011, ch. 6; and though limited only to stimuli near the center of the visual field, see the large literature on "overflow" in response to Block 2007.

This is sometimes called the “refrigerator light illusion”, akin to the error of thinking the refrigerator light is always on because it’s always on when you open the door to check.[107] On this view, you’re only tempted to think you have constant tactile experience of your feet in your shoes because you have that experience on those rare occasions when you’re thinking about whether you have it. Even if you now seem to have a broad range of experiences in different sensory modalities simultaneously, this could result from an unusual act of dispersed attention, or from “gist” perception or “ensemble” perception, in which you are conscious of the general gist or general features of a scene, knowing that there *are* details, without actually experiencing those unattended details.[108]

The opposite mistake is also possible. Those who deny a constant stream of peripheral experiences might simply be failing to notice or remember them. The fact that you don’t remember *now* the sensation of your feet in your shoes two minutes ago hardly establishes that you lacked the sensation at the time.

Although many people find it introspectively compelling that their experience is rich with detail or that it is not, the issue is methodologically complex because introspection and memory are not independent of the phenomena to be observed.[109]

If we do have rich sensory experience outside of attention, it is unlikely that all of that experience is present in or broadcast to a Global Workspace. Unattended peripheral information is rarely remembered or consciously acted upon, tending to exert limited downstream influence – the paradigm of information that is *not* widely broadcast. Moreover, the Global Workspace is generally characterized as limited capacity, containing a very limited number of thoughts, representations, objects, or processes at any one time – those that survive some competition or attentional selection – not a welter of richly detailed experiences in many modalities at once.[110]

[107] Thomas 1999.

[108] Oliva and Terralba 2006; Whitney and Leib 2018.

[109] Schwitzgebel 2007 explores the methodological challenges in detail.

A less common but equally important objection runs in the opposite direction: Perhaps not everything in the Global Workspace is conscious. Some thoughts, representations, or processes might be widely broadcast, shaping diverse processes, without ever reaching explicit awareness.[111] Implicit racist assumptions, for example, might influence your mood, actions, facial expressions, and verbal expressions. The goal of impressing your colleagues during a talk might have pervasive downstream effects without occupying your conscious experience moment to moment.

The Global Workspace theorist who wants to allow that such processes are not conscious might suggest that not all widely broadcast processes occupy the Global Workspace. This then raises the challenge of distinguishing those that do from those that don't. One possibility: In adult humans, processes in the workspace are generally also available for introspection (one type of luminosity).[112] However, if the workspace is redefined primarily in terms of introspectibility, this amounts to shifting to a Higher Order view.

*3. Generalizing Beyond Vertebrates.*

The empirical questions are difficult even in ordinary adult humans. But our topic isn't ordinary adult humans – it's AI systems. For Global Workspace Theory to deliver the right answers about AI consciousness, it must be a *universal* theory applicable everywhere, not just a theory of how consciousness works in adult humans, vertebrates, or even all animals.

[110] E.g., Dehaene 2014; Mashour et al. 2020.

[111] E.g., Searle 1983, ch. 5; Bargh and Morsella 2008; Lau 2022; Michel et al. 2025.

[112] E.g., Shea and Frith 2019, though their view is expressed in terms of metacognition generally rather than introspection specifically.

If there were a sound *conceptual* argument for Global Workspace Theory, then we could know the theory to be universally true of all conscious entities. Empirical evidence would be unnecessary. It would be as inevitably true as that rectangles have four sides. But as I argued in Chapter Four, conceptual arguments for the essentiality of any of the ten possibly essential features are unlikely to succeed – and a conceptual argument for Global Workspace Theory would be tantamount to a conceptual argument for the essentiality of access, one of those ten features. Not only do the general observations of Chapter Four suggest against a conceptual guarantee, so also does the apparent conceivability, as described in Section 2 of this chapter, of consciousness outside the workspace or nonconsciousness within it – even if such claims are empirically false.

Consequently, if Global Workspace Theory is the correct universal theory of consciousness applying to all possible entities, an empirical argument must establish that fact. But it's hard to see how such an empirical argument could proceed. We face another version of the Problem of the Narrow Evidence Base. Even if we establish that in ordinary humans, or even in all vertebrates, a thought, representation, or process is conscious if and only if it occupies a Global Workspace, what besides a conceptual argument would justify treating this as a universal truth that holds among all possible conscious systems?

Consider some alternative architectures. The cognitive processes and neural systems of octopuses, for example, are distributed across their bodies, often operating substantially independently rather than reliably converging into a shared center.[113] AI systems certainly can be, indeed often are, similarly decentralized. Imagine coupling such disunity with the capacity for self-report – an animal or AI system with processes that are reportable but poorly integrated with other processes. If we assume Global Workspace Theory at the outset, we can conclude that only sufficiently integrated processes are conscious.[114] But if we don't assume Global Workspace Theory at the outset, it's difficult to imagine what near-future evidence could establish that fact beyond a reasonable standard of doubt to a researcher who is initially drawn to a different theory.

[113] Godfrey-Smith 2016b; Carls-Diamante 2022.

If the simplest version of Global Workspace Theory is correct, we can easily create a conscious machine. This is what Dehaene, Lau, and Kouider envision in the 2017 paper I discussed in Chapter One. Simply create a machine – such as an autonomous vehicle – with several input modules, several output modules, a memory store, and a central hub for access and integration across the modules. Such a system meets the minimal criteria of Global Workspace Theory, with the hub as the workspace. Consciousness then follows. If this seems doubtful to you, then you cannot straightforwardly accept the simplest version of Global Workspace Theory.[115]

---

[114] I further explore partly disunified consciousness, group consciousness, and borderline cases that defy discrete assessment in Schwitzgebel 2023; Schwitzgebel and Nelson 2023, 2025.

[115] One might alternatively read Dehaene, Lau, and Kouider 2017 purely as a conceptual argument: *If* all we mean by "conscious" is "accessible in a Global Workspace", *then* building a system of this sort suffices for building a conscious entity. The difficulty then arises in moving from that stipulative conceptual claim to the interesting, substantive claim about phenomenal consciousness in the standard sense described in Chapter Two. Similar remarks apply to the Higher Order aspect of that article. One challenge for this deflationary interpretation is that in related works (Dehaene 2014; Lau 2022) the authors treat their accounts as accounts of phenomenal consciousness. The 2017 article also concludes by emphasizing that in humans "subjective experience coheres with possession" of the functional features they identify. A further complication: Lau later says that the way he expressed his view in this 2017 article was "unsatisfactory": Lau 2022, p. 168. See also Goldstein and Kirk-Giannini (forthcoming) for an extended application of Global Workspace Theory to AI consciousness.

We can apply Global Workspace Theory to settle the question of AI consciousness only if we know the theory to be true either on conceptual grounds or because it is empirically well established as the correct universal theory of consciousness applicable to all types of entity. Despite the substantial appeal of Global Workspace Theory, we cannot know it to be true by either route.

*4. Higher Order Theories and Luminosity.*

Among the main competitors to Global Workspace theories are Higher Order theories. Where Global Workspace theories treat access as the central essential feature of consciousness, Higher Order theories traditionally privilege luminosity.[116] Luminosity – recall from Chapter Three – is the thesis that having an experience entails knowing about that experience or at least being in a position to know about it, or that conscious experiences are inherently self-representational, or that having an experience entails being in some sense aware of it. As noted in Chapter Four, Higher Order Theories can be motivated by a seeming-tautology: To be in a conscious state is to be conscious *of* that state, which requires representing it in a certain way.[117] This is not actually a tautology, but a substantive claim. *Maybe* consciousness requires representing one's own mental states, but if so, that is a nonobvious fact about the world, not a simple conceptual truth.

[116] One important exception is Brown 2025, who advocates a non-traditional Higher Order Theory that is not committed to luminosity. As he argues, cases of "radical misrepresentation" of the target state create a challenge for traditional Higher Order views. Also, if introspection requires *third*-order representation (Rosenthal 2005), then a merely second-order state might potentially make a first-order state conscious, and something you are in one sense luminously "aware of", even without introspectibility.

[117] E.g., Lycan 2001; Rosenthal 2005.

Like Global Workspace Theory, Higher Order theories have some initial appeal. In the typical adult human case, when we have conscious experiences we seemingly have some knowledge of or awareness of them – perhaps indirect, inchoate, and not explicitly conceptualized.[118] This needn't imply infallibility. When we attempt to describe or categorize that experience, we might err. Consider again some typical experiences: your visual experience of this text, a sting of pain, a tune in your head, that familiar burst of joy when you see a cute garden snail. Plausibly, as they occur, you know they are occurring – or if "knowledge" is too strong, at least you have some acquaintance with them, some attunement or potential attunement to the fact that they are going on.[119]

Also like Global Workspace Theory, Higher Order theories can potentially explain other possibly essential features of consciousness. Maybe the relevant type of self-representation or self-awareness entails experiencing a self, or a subject. If so, *subjectivity* follows. Maybe self-representation or self-awareness is only possible if the thought, representation, or process is also available for other types of downstream cognition – or maybe the higher order representation serves as a gatekeeper for other downstream processes. If so, *access* follows. Maybe there's always a determinate fact about whether a thought, process, or representation is or is not targeted by a higher order process or representation, which could potentially explain *determinacy.* If the represented states are themselves always representations, and if all representations are necessarily about something, that could explain *intentionality*.

[118] See also views inspired by Brentano 1874/1973, such as Kriegel 2009; Spener 2024.

[119] On acquaintance theories, see Gertler 2010, ch. 4; Giustina 2022; Duncan forthcoming.

And just as Global Workspace Theories suggest an architecture for AI consciousness, so also do Higher Order Theories suggest an architecture, or at least a piece of an architecture: Any conscious system must monitor its own cognitive processing.

For this architectural interpretation, a challenge immediately arises: the Problem of Minimal Instantiation.[120] This problem arises for most functionalist theories of consciousness – compare Dehaene, Lau, and Kouider's self-driving car – but it's especially acute here. Any machine that can read the contents of its own registers and memory stores can arguably, in some sense, represent its own cognitive processing. If this counts as higher order representation and if higher order representation suffices for consciousness, then most of our computers are already conscious!

A Higher Order Theorist can resist this radical implication in at least three ways: (1) by denying that this is the right kind of self-representational process (opening the question of what the right kind is); (2) by denying that the lower-order processes are the right kind of targets (perhaps they are not genuine thoughts or representations); (3) or by requiring some further necessary condition(s) for consciousness. Alternatively, the Higher Order Theorist can accept that consciousness is much more widespread than generally assumed.

Higher Order theories differ in flavor. On Lau's Perceptual Monitoring Theory, representations become conscious when a discriminator mechanism judges a sensory representation not to be random "noise" and makes it available for downstream cognition. These representations needn't be *globally* broadcast, as long as they have "an appropriate impact on a narrative system capable of causal reasoning".[121] On Lau's view, constructing a conscious robot would be fairly straightforward. It might, for example, have cameras that generate computational representations of its environment, an ability to assess how similar or dissimilar those representations are to each other, and the ability to assess the likelihood of error under various conditions.[122]

---

[120] Compare Herzog et al. 2007; Butlin et al. forthcoming; and Goldstein and Kirk-Giannini forthcoming on "small models".
[121] Lau 2022, p. 209.

Axel Cleeremans' Self-Organizing Metarepresentational Account demands much more. On this view, consciousness arises when a system representationally redescribes its inner workings to better predict the consequences of its actions in the world, especially in social contexts where it learns to represent itself as one agent among others. We "learn to be conscious" when we build models of the internal, unobservable states of agents in the world like ourselves.[123] The required social modeling appears to be well beyond the capacity of all but the most socially sophisticated animals, suggesting that consciousness will be sparsely distributed in the animal kingdom. However, nothing in the theory suggests that a sophisticated, embodied, socially embedded AI system would be incapable of achieving the right types of higher order representation.

Non-traditional Higher Order theories de-emphasize luminosity. On Richard Brown's Higher Order Representation of a Representation account, the lower-order target representation needn't even exist.[124] On Michael Graziano's Attention Schema Theory, what we think of as consciousness is just a simplified model of our attentional processes.[125] We can't explore the details here, but the unifying feature is that consciousness depends on representing one's own mind in a particular way.

[122] Lau 2022, p. 211-212. Regarding non-human animals, Lau's view is surprisingly restrictive, holding that many smaller mammals likely have no conscious experiences, p. 167.

[123] Cleeremans et al. 2020; Fleming, Brown, and Cleeremans forthcoming.

[124] Brown 2025.

[125] Graziano 2019.

*5. Consciousness Without Higher Order Representations; Higher Order Representations Without Consciousness?*

Could some states be consciously experienced without being targeted by higher order representations? Rich, unattended sensory experiences – if they exist – again pose a challenge. Higher order representations that duplicate the finely detailed content of lower order representations would seemingly clutter the mind with needless redundancy. More plausibly, higher order representations might encode gist or ensemble summary content ("lots of red dots over there"), omitting the individual details from experience.[126]

The Sampling Bias problem (from Chapter Four) also arises: Introspecting and recalling experience might require higher order representations, and thus all the experiences *you know about and report* might involve them, but that doesn't entail that *all of your experiences full stop* involve higher order representations. At least in principle, it seems that you might have many unrepresented and unremembered experiences. Theories that liberally ascribe consciousness to nonhuman animals – such as Integrated Information Theory, Recurrence Theories, and Associative Learning Theories (see Chapter Nine) – support this possibility. All appear to allow that the right informational or cognitive complexity might generate experience without higher order representation. If an ant or snail might have conscious experiences that aren't targeted by higher order representations, so also sometimes might you.

[126] Brown 2012, 2025; Mudrik et al. 2025.

Conversely, might some cognitive processes be targeted by higher order representations but not consciously experienced? Research in metacognition suggests that the mind keeps constant tabs on itself. The ordinary flow of speech requires that we track a huge amount of information about background assumptions we share with our interlocutors, the logical implications and pragmatic implicatures of our and others' utterances, and what contextual information we should provide to facilitate our partner's understanding. Tracking all of this arguably involves considerable self-representation – of your aims, of what your partner knows about your aims, and of what you and your partner know in common.[127] Intentional learning (e.g., studying for a test) requires constantly assessing the shape of your knowledge and ignorance, where to most profitably focus attention, when to start and stop, and the likelihood of later recognition or recall. It's doubtful that all of these metacognitive judgments generate conscious experience of the lower order states they are responding to. Ordinary motor activities arguably require metarepresentationally tracking progress toward goals and the potential success or failure of subplans, adjusting movement on the fly at a pace and with a degree of detail that we ordinarily think of as outside of conscious awareness.[128] A Higher Order Theorist can deny that these are the right types of higher order representation, or that they involve higher order representation at all, but that creates the challenge of explaining what's in and what's out. If many nonconscious processes meet the structural criteria for higher order representation, the theory must supply principled grounds for their exclusion.

*6. Generalizing Beyond Vertebrates Again.*

[127] Wilson and Sperber 2012; Brown-Schmidt and Heller 2018.
[128] Gallivan et al. 2018; Christensen et al. 2019.

Recent Higher Order Theorists rightly treat the theory not as a conceptual truth but as an empirical hypothesis with testable implications. Consequently, even if *in humans* higher order representations of the right sort are both necessary and sufficient for consciousness, the extension to animal, alien, and AI cases is conjectural rather than conceptually guaranteed.

To illustrate the types of consideration invoked: Lau argues that Higher Order Theory has an empirical advantage over Global Workspace Theory because nonconscious sensory stimuli sometimes influence a wide range of cognitive processes, suggesting that nonconscious representations can occupy the "workspace", contra Global Workspace Theory.[129] Lau also argues against "local" theories (such as Recurrence Theory, Chapter Nine) that invisible stimuli activate the visual cortex as much as visible stimuli do, once task performance capacity is properly controlled for, and consequently some further downstream processing, not just local processing in the visual cortex, must be necessary for consciousness.[130] (One example of an "invisible" stimulus would be an image so effectively masked by flickering lights that participants report not having seen it.) Rejoinders are possible, and the science is uncertain and evolving. Lau himself emphasizes the difficulty of directly testing the hypotheses at issue.[131]

The crucial experiments are conducted in humans or other vertebrates such as monkeys, which returns us to the Problem of the Narrow Evidence Base (Chapter Four and Section 3 of this chapter). If Higher Order Theory is an empirical hypothesis, generalizing from vertebrates to a universal claim that applies to AI systems requires a huge speculative extrapolation. Something more might be needed in addition to higher order representations, undermining the sufficiency of higher order representations for consciousness – a background (e.g., biological) condition met in humans but not in AI systems.  Alternatively, undermining the necessity of higher order representations, even if our lovely primate way of generating consciousness always involves them, other (e.g., simpler) entities might generate consciousness differently.

[129] Lau 2022, 129-132.

[130] Lau 2022, 132-135.

[131] Lau 2022, 2025.

*7. Close Kin.*

Global Workspace Theories and traditional Higher Order Theories are close kin. Thoughts, processes, or representations are conscious if they influence later cognition in a particular way, either through becoming broadly accessible across the cognitive system or being targeted by a further representational process.  If broad access and higher order representation are closely linked, with one typically enabling the other, each approach can to a substantial extent explain the other's successes, making them challenging to empirically distinguish.

Both theories are most naturally interpreted as suggesting that people don't experience a rich welter of simultaneous experiences in many modalities – an advantage if the contents of experience are relatively sparse, a liability if experience is in fact rich.  Both theories draw their empirical support from human and other vertebrate cases, leaving unclear how far we can extrapolate to very different types of systems, such as AI.  And both are potentially vulnerable to the Problem of Minimal Instantiation: It seems easy to create simple AI systems that meet the minimal criteria of these theories but which most theorists would hesitate to regard as conscious.

Despite these similarities, their implications for AI consciousness are very different, since it seems eminently possible to create an AI system with a global workspace but no higher order representation or vice versa.

## Chapter Nine: Integrated Information, Local Recurrence, Associative Learning, and Iterative Natural Kinds

This chapter examines three other prominent theories, each quite different from the theories of Chapter Eight as well as from each other: Integrated Information Theory, Local Recurrence Theory, and Unlimited Associative Learning. It concludes with reflections on the possibility of less theory-laden empirical approaches.

### *1. Integrated Information Theory.*

Integrated Information Theory (IIT), developed by Guilio Tononi and collaborators, treats consciousness as a matter of (you guessed it) the integration of information. In IIT, "information" is understood as causal influence, mathematically formalized.[132] Systems with high information integration typically feature complex, looping causal processes, specialized subunits, and dense interconnectivity. The idea that the human brain's complex information management explains its high degree of consciousness has both empirical and intuitive appeal. It would also delight a certain clade of nerds (I am one) to satisfactorily explain consciousness through a rigorous mathematical formalism grounded in objective facts about causal connectivity.

Empirical measures of "perturbational complexity" provide some support for Integrated Information Theory. Perturbational complexity is typically measured by disturbing the brain with outputs from an electromagnetic coil and assessing the complexity of subsequent electrical (EEG) scalp recordings. Responses to electromagnetic perturbation is generally more complex when subjects are in highly conscious states – wakefulness and sleep phases associated with

[132] Albantakis et al. 2023.

dreaming – than in coma and sleep phases associated with less dreaming.[133] Neurophysiological architecture offers further support for IIT: The cortex, or the cortex plus thalamus and other related areas, shows more connective complexity than the cerebellum. As IIT predicts, cortical disorders tend to affect conscious experience much more than cerebellar disorders.

Despite its appeal, Integrated Information Theory faces serious challenges. It proposes a general measure of information integration, Φ, which is computationally intractable for most systems, making the theory difficult to rigorously test.[134] The theory has unintuitive consequences, such as attributing a small amount of consciousness to some tiny feedback networks and potentially superhuman degrees of consciousness to some large but simply-structured networks, provided their components are linked in the right way.[135] Where calculations of Φ are tractable, results often fluctuate dramatically with small changes in connectivity, in contrast with the robustness of the human brain.[136] And standard versions of IIT hold that subsystems cannot be conscious if they are embedded in larger more informationally complex systems, meaning that consciousness does not depend only on local processes. This leads to the counterintuitive result that arbitrarily large amounts of consciousness can appear or vanish with the loss or addition of a single bit of information in a system's surroundings, even if that information is not currently influencing the system's internal operations.[137] Advocates of

---

[133] Casali et al. 2013; Casarotto et al. 2016.

[134] Mediano et al. 2022; for one prominent attempt not requiring precise calculation of Φ, see Cogitate Consortium et al. 2025.

[135] Aaronson 2014.

[136] Mediano et al. 2018; Schwitzgebel 2018.

[137] Schwitzgebel 2014.

Integrated Information Theory swim gleefully against this tide of troubles; the theory is not decisively refuted and remains influential.

Since it requires neither a global workspace in the traditional sense nor higher order representations, Integrated Information Theory has very different implications for what AI systems would be conscious, and to what degree, than do Global Workspace Theory and Higher Order Theory. For example, some systems (e.g., a network of the right kind of logic gates in a two-dimensional grid) would be conscious on IIT (to an arbitrarily high degree depending on the size of the network[138]) but not on most other views. Conversely, a computer designed to implement a Global Workspace or employ Higher Order representations might, according to IIT, have consciousness only in some subsystems but not as a whole: Standard computers are engineered for modular decomposability, with the elements designed to operate as separate components, which then deliver results to each other as discrete outputs. Most of the information integration is within subcomponents which communicate with each other, rather than through globally integrated causal processes across the whole. Consciousness will arise in whichever components integrate the most information as measured by $\Phi$, but these will be multiple smallish-$\Phi$ subsystems with limited consciousness. There will be no humanlike degree of consciousness across the system as a whole.[139]

*2. Local Recurrence Theory.*

As noted in Chapter Eight, Global Workspace Theory and Higher Order theories invite the idea that sensory experience is sparse. Only what is selected to enter a relatively restricted workspace for broadcast across the mind, or only what is targeted by (presumably selective) higher order representations, becomes conscious.

[138] Aaronson 2014.

[139] Findlay et al. 2024/2025.

Local Recurrence Theory, developed by Victor Lamme, denies that such further "downstream" processing is necessary.[140] In vision, signals from the retina travel quickly to the occipital cortex at the back of the brain, where specialized neurons react selectively to features like motion, color, and edges at various orientations. These neurons then send signals forward to frontal, temporal, and parietal regions. Global Workspace and Higher Order theories typically require such downstream signaling for consciousness: the workspace and the processes underlying higher order representation are assumed not to reside in the occipital cortex. Local Recurrence Theory, in contrast, holds that the right kind of local processing in "early" occipital regions can generate consciousness on its own. However, not just any activation of early sensory regions will do. There must be sufficient *recurrent* processing – signals must interact in causal loops, integrating perceptual information. For example, occipital processes responsive to color might influence occipital processes responsive to shape and vice versa, creating a unified perceptual experience of a red square. The common impression that experience is rich with detail in many sensory modalities at once can then be preserved, as long as the right type of recurrent processing occurs in each modality. The downstream processes described by Global Workspace Theory and Higher Order theories might be necessary for a *reportable* and *memorable* perceptual experience, and for broad accessibility of perceptual information, but on Local Recurrence Theory such access is not necessary for consciousness.

In principle, the dispute between local/early theories, like Local Recurrence Theory, and global/late theories, like Global Workspace and Higher Order theories, is empirically testable.

---

[140] Lamme 2006, 2010.

For example, researchers can examine cases where early neural areas are highly active while later areas are not, and vice versa, to see which pattern correlates better with reports of consciousness. In practice, however, empirical adjudication is difficult. Confronted with high levels of neural activity in early areas but no reports of consciousness, local theorists can suggest that the activity is insufficiently recurrent or that the activity is conscious but – just as their theory would predict – unreportable because it is inadequately processed further downstream. Also, in ordinary, intact brains, local activity tends to have downstream consequences, and downstream activity tends to influence upstream areas. Untangling these effects is difficult given the limitations of current neuroimaging techniques. The empirical debates continue, with some results more easily accommodated on local theories and other results more easily accommodated on global or higher order theories. No decisive resolution is likely in the near-to-medium term.[141]

But let's not lose sight of our particular target: consciousness or its absence in AI systems. Suppose Local Recurrence Theory eventually prevails. In humans, and maybe in all vertebrates, recurrent loops of local sensory processing are both necessary and sufficient for consciousness. Would this generalize to AI systems? Recurrent loops of processing are common in AI – even in simple systems. The Problem of Minimal Instantiation thus arises again. Is my laptop conscious every time it executes a recurrent function or integrates information in causal loops? Presumably, consciousness requires *enough* recurrence, of the right *type*, and perhaps in a context of background conditions we take for granted in humans but which might not exist in artificial systems.

[141] Phillips 2018; Block 2023; Lau 2025.

Extending Local Recurrence Theory to AI would require deeper reflection on what makes recurrence the right kind of process to generate consciousness. One natural answer appeals to *unity* as an essential feature of consciousness. A frequently suggested role for recurrence is in binding together visual features that are registered by different clusters of neurons (such as color and shape). The importance of recurrence in conscious primate vision might then derive from its importance in generating a unified perceptual experience.[142]

Unlike Integrated Information Theory, Local Recurrence Theory is not typically framed as a universal theory of consciousness applicable to all possible entities, whether human, animal, alien, or AI. It's an empirical conjecture about human consciousness, drawing mainly on studies of human and monkey vision. Substantial theoretical development and speculation would be needed to adapt it to AI cases. Still, recognizing it as a live competitor to Global Workspace and Higher Order theories highlights the diversity of theories of human consciousness – how far we remain from a good understanding of the basis of consciousness even in our favorite animal.

*3. Unlimited Associative Learning.*

Another influential theory – the last we will consider – is Simona Ginsburg's and Eva Jablonka's Unlimited Associative Learning, which holds that consciousness arises when a particular kind of cognitive capacity is present.[143]

Ginsburg and Jablonka begin with a list of seven attributes of conscious experience, which they derive from an overview of the scientific and philosophical literature – attributes, they suggest, that are "individually necessary and jointly sufficient for consciousness".

---

[142] Lamme 2010; Roelfsema 2023.

[143] Ginsburg and Jablonka 2019.

1. *Global activity and accessibility.* Conscious information is not confined to one region but globally available to cognitive processes.
2. *Binding and unification.* Features of experience, such as colors and shapes, sights and sounds, are integrated into a unified whole.
3. *Selection, plasticity, learning, attention.* Conscious experience involves "the perception of one item at a time"; and it involves neural and behavioral adaptability to changing circumstances.
4. *Intentionality.* Conscious states represent and are "about" things.
5. *Temporal thickness.* Consciousness persists over time, due to recurrent processes, reverberatory loops, and the activation of networks at several scales.
6. *Values, emotions, goals.* Experiences have subjective valence, feeling positive or negative.
7. *Embodiment, agency, and self.* Consciousness involves a stable distinction between one's body and the environment, plus a feeling of ownership or agency.[144]

Even the sleepy reader will notice a resemblance between these seven features and the ten possibly essential features of consciousness described in Chapter Three.[145]

Ginsburg and Jablonka draw on a wide range of animal studies suggesting that the animals whose cognition manifests these seven features also exhibit "unlimited associative learning". Unlimited associative learning is best understood by contrasting it with the limited associative learning of cognitively simpler animals like the *C. elegans* nematode worm and the

[144] Ginsburg and Jablonka 2019, p. 98-101

[145] One contrast: Neither valence nor learning/plasticity appear on Chapter Three's list of ten. As criteria for necessary features of particular, individual experiences, they seem too demanding: Plausibly, individual experiences can be emotionally neutral and leave no enduring mark. Perhaps, however, species who are conscious must have some experiences with these features.

*Aplysia californica* sea hare. *C. elegans* and *Aplysia californica* can learn to associate a limited range of stimuli with stereotypical responses. For example, sea hares learn to withdraw their gills when gently prodded if the prod is repeatedly paired with a shock. In contrast, animals capable of unlimited associative learning – many or all vertebrates and arthropods (insects and crustaceans), plus the more cognitively sophisticated mollusks (such as the octopus) – can learn complex behavioral adjustments to a wide range of complex stimuli. Octopuses can learn to unscrew jars to get food; rats can learn complex mazes; bees can learn to pull on string to retrieve drops of sucrose solution from behind Plexiglas – and can even learn socially by watching each other.[146]

Ginsburg and Jablonka acknowledge that consciousness might exist in animals with only limited associated learning, who exhibit some but not all of the seven features.[147] More relevant to our topic, they allow that unlimited associative learning in a robot might be insufficient for consciousness, if the robot lacks some other essential biological features (which they don't further specify).[148] Still, if there's a division in nature between animals with and without the capacity for unlimited associative learning, and if that division corresponds with the seven features Ginsburg and Jablonka attribute to consciousness, then, speculatively, the capacity for unlimited associative learning might mark the dividing line between animals that are and are not conscious, and – even more speculatively – AI consciousness might require the same capacities.

Whether the seven features listed by Ginsburg and Jablonka are indeed all necessary for consciousness is an open question. We've just seen one theory – Local Recurrence Theory – that denies the necessity of downstream accessibility. And perhaps we can imagine weird alien or AI systems who are conscious but who lack one or more of the other seven features. Alternatively, Ginsburg's and Jablonka's list might omit some essential feature. Higher Order theories hold that higher order representations of one's mentality are necessary for consciousness. Even if we accept the list of seven, substantial further research will be needed to establish the tight connection between unlimited associative learning and these features, and exactly what kinds of accessibility, binding, plasticity, etc., are required, and how to generalize from animals to AI.

---

[146] Chittka 2022.

[147] Ginsburg and Jablonka 2019, p. 395.

[148] Ginsburg and Jablonka 2019, p. 227, 395.

*4. General Observations about Theory-Driven Approaches.*

If we had the right universal theory of consciousness, we could apply it to AI systems to determine whether they are conscious. Problem solved! I've offered a selective tour of some currently prominent candidate theories.[149] What I hope this tour suggests is:

First, there is no consensus on a general theory of consciousness even for the human case, nor is such consensus likely anytime soon.

Second, apart from Integrated Information Theory, it's unclear how to apply these frameworks to AI. How much information sharing is enough? What type and degree of recurrence? What kinds of self-representation? Are biological conditions needed in addition to associative learning? Most theories face the Problem of Minimal Instantiation: Tiny AI implementations seem to meet their criteria for consciousness, in systems that people would generally regard as nonconscious.

Third, to the extent these theories are empirical, they face the Problem of the Narrow Evidence Base. Suppose – very optimistically! – that over the next several decades scientists converge on a consensus theory of human consciousness, or vertebrate consciousness, or even consciousness in all animals on Earth. AI systems differ radically in structure. Applying theories developed for animals to such alien architectures might be like applying a theory of animal biology to a computer chip. It's a huge extrapolatory leap. If there were a sound, purely conceptual argument that all conscious systems have such-and-such features, we could look for those features in AI. But if the arguments are empirical, grounded in animal cases, it's difficult to see how to bridge from our knowledge of animals to artificial systems.

[149] For longer lists, see Seth and Bayne 2022; Kuhn and Gómez-Marín 2025/2026.

Although this reasoning does not reduce entirely to the argument at the end of Chapter Three, it can be cast in those terms. The wide range of viable scientific theories leaves us justifiably unsure which among the ten possibly essential features of consciousness is truly essential. If we cannot at least address that basic question, we will remain in the dark about the consciousness of near-future AI. Uncertainty about the features reinforces uncertainty about the theories; uncertainty about the theories reinforces uncertainty about the features.

*5. Iterative Natural Kinds.*

Despite these pessimistic reflections, the darkness is not pitch. One partly hopeful thought is this: By conjoining the features of plausible theories, we can reach tentative judgments about the *relative* likelihood of the consciousness, or not, of different AI systems. Unless you have strictly zero credence in the possibility of AI consciousness, or zero credence that any of the leading theories point in approximately the right direction, you should allow that a system with all the features favored by those theories is likelier to be conscious than a system with none of the features. Suppose, for example, that an AI system develops in a biological substrate, with a neuromorphic structure and a single global workspace where information is integrated and broadcast downstream, in complex causal processes that cannot easily be informationally compressed, with plenty of recurrent processing, the capacity for sophisticated, flexible responses to challenging real-world environments, unlimited associative learning, self-representation, and accurate verbal self-reports. Add further features if you like. Such a system is more plausibly conscious than a system with none of those features. It might still be reasonable to doubt its consciousness, perhaps even to give it much less than a 50% chance of being conscious – but such a system would be better hunting grounds than a mimicry-based language model. Patrick Butlin, Robert Long, and their collaborators have called this the Indicator Properties strategy for evaluating the potential consciousness of AI systems.[150]

Another hopeful thought is inspired by analogy to breakthroughs in measurement science. In his influential treatment of the history of thermometry, Hasok Chang confronts what seems to be a methodological paradox.[151] How do you calibrate the first thermometer? Calibrating a thermometer seems to require a more accurate thermometer – but none yet exists. Alternatively, you might appeal to a good theory of temperature – but that doesn't yet exist either, not without an accurate thermometer against which the theory can be tested. The solution was to advance gradually by baby steps from rough, intuitive measures to more rigorous ones. For example, sensations of hot or cold can be correlated with the expansion and contraction of fluids. Fluid expansion and contraction can then be used to correct sensations, especially when there's reason to think the sensations might be misleading (e.g., a lukewarm object feeling cold to a hand previously immersed in warm water) and when touch is impractical (e.g., with very hot objects). The problem of measurement isn't immediately solved, since different fluids expand in different patterns, and fluids are held in measuring containers that also frustratingly expand, and solid objects and gases also have temperatures…. However, by correlating enough tests, and using them to correct each other especially when one test might be better than another for a particular circumstance, scientists eventually converged on highly accurate thermometers and a well-founded theory of temperature.

[150] See Butlin et al. 2023; Butlin et al. forthcoming.

[151] Chang 2004.

Inspired in part by this example, some researchers – for example, Tim Bayne, Nicholas Shea, and Andy McKilliam – suggest that consciousness science can advance similarly, despite lacking consensus measures and theories.[152] A first step might be noticing behavioral and neurophysiological correlates of consciousness in typical adult humans. One behavioral candidate is trace conditioning – the capacity to learn an association between two stimuli across a temporal gap. It has been argued that in humans this is possible only when the stimuli are consciously perceived.[153] One neural candidate is widespread neural activity about 300 milliseconds after the onset of a stimulus.[154] Such measures might be used to correct introspective reports, especially if there's reason to think the introspections might be inaccurate, and to measure consciousness when introspective report is impossible, for example, expanding the measure to other primates. Adjust and expand, adjust and expand, adjust and expand… and eventually, *maybe* a diversity of measures will converge toward the same results, each compensating for the others' weaknesses. We can then claim to have accurately measured consciousness, and we can build our theory accordingly. This is sometimes called the *iterative*

---

[152] Bayne and Shea 2020; Bayne et al. 2024; McKilliam 2025.

[153] Birch 2022; perhaps especially if it is susceptible to failure absent attention: Droege et al. 2021. But see Michel forthcoming.

[154] Dehaene 2014.

*natural kind* strategy, since it assumes that consciousness is a "natural kind" like gold, water, or kinetic energy, around which scientific regularities congregate.

This strategy will fail if consciousness is a loose amalgam of several features or if it splinters into multiple distinct kinds. But even such failures could be informative. We might discover that phenomenal consciousness – what-it's-like-ness, experientiality – is not one thing but several related things or a mix of things, much as we learned that "air" is not one thing. In the long term, it's not unreasonable to hope for either convergence toward a single natural kind or an informative failure to converge. However, this is a much longer-term prospect than the development of AI systems that a significant portion of experts and ordinary people are tempted to regard as conscious. I don't claim that it's impossible to develop a scientifically well justified universal theory of consciousness that applies to all possible creatures, whether human, animal, alien, or AI. But it's a distant hope.

## Chapter Ten: Does Biological Substrate Matter?

So far, every entity that is generally recognized to be conscious is biological. Maybe some biological property is crucial? If so, and if no near-future AI could replicate that property, we could dismiss the possibility of AI consciousness without worrying about the theoretical or empirical details.

The first section of this chapter will discuss one such candidate property: autopoiesis.

The second section will discuss and reject one prominent critique of biological views: the neural replacement argument.

The third section will offer a "Copernican" argument that consciousness should not require similarity to us in fine-grained biological detail, which suggests at least some flexibility in the substrate of consciousness.

### *1. Autopoiesis.*

The idea of *autopoiesis* was introduced by Humberto Maturana and Francisco Varela in 1972.[155] Autopoietic (self-creating) entities continuously regenerate their own components, maintain their structure and processes over time, and constitute themselves as distinct from their environment. Living organisms are autopoietic: They synthesize their constituent molecules, draw energy from outside to maintain homeostasis, and protect themselves with skins, shells, walls, and membranes. Philosopher Evan Thompson and neuroscientist Anil Seth have argued that consciousness requires autopoiesis of the sort we do not see in AI systems.[156] Perhaps this is the most prominent argument for biologicism about consciousness.

[155] Maturana and Varela 1972/1980.

Autopoiesis establishes a boundary between self and other – an aspect of *subjectivity,* one of the possibly essential features of consciousness discussed in Chapter Three. Autopoiesis also suggests norms and purpose. Things can go well or poorly for autopoietic systems. A well-functioning autopoietic system is also a unity, harmoniously maintaining itself. Sufficiently sophisticated and self-protective autopoietic systems might also exhibit privacy, flexible integration, and access – perhaps also self-representation and a sense of the present versus past and future. Living, autopoietic systems can be just the sorts of things to manifest features that we normally associate with consciousness, including some of the ten possibly essential features from Chapter Three. Following Thompson and Seth, one might then hold that (1) autopoiesis is necessary for consciousness, and (2) no near-future AI could be autopoietic. There's an aesthetic appeal, too, in linking together arguably the two most special features of Earth, life and mind; the view sparkles with *je ne sais quoi*.

However, assertion (1) requires justification, and prominent autopoietic theories generally highlight the attractions of a link between autopoiesis and consciousness without presenting any sustained explanation of why non-autopoietic systems couldn't also be conscious. Life is great! But perhaps non-life can also be great, at least in the respects necessary for consciousness. The claim that *only* autopoietic systems can be conscious lacks a well-developed theoretical defense.

In any case, contra assertion (2), AI systems can plausibly be autopoietic. Think beyond desktop computers and language models stored in the cloud. For example: A solar-powered robot might seek energy sources. It might have error checking programs that detect and discard defective parts. It might build new parts from local materials or order components online and assemble them for self-repair. It might detect and reject fake parts and repel intrusive materials. It might be composed of modules held together electromagnetically so that maintaining itself as a coherent whole requires it to constantly expend energy. A plastic shell might maintain the boundary between the robot and its surroundings. Internal sensors might detect flaws in its shell by detecting light through unexpected cracks, by visually monitoring its exterior, and by electrostatically detecting gaps. Perhaps its shell coating is continuously refurbished by capillaries that emit lacquer as needed. If the robot has sufficient redundancy, it could also detect flaws in and replace central processing systems. The robot might even manufacture duplicates or near-duplicates of itself with the same capacities, creating an evolutionary lineage. While such a system would lack the rich multi-level autopoiesis of living systems and wouldn't constantly manufacture its own parts at the chemical level, it appears to meet theoretical minimal criteria for autopoiesis.[157]

---

[156] Thompson 2007; Seth forthcoming. Godfrey-Smith's (2016a) suggestion that metabolism is what's crucial faces challenges similar to those I pose for autopoiesis in this section. First, it's unclear why it should be so important, and second, AI systems might have metabolisms.

In AI technologies more directly modeled on life – artificial life systems or DNA-based computing – the autopoietic features potentially become richer. Although such systems have not been developed or deployed on anything like the scale of standard computer-based systems, they do exist and could potentially become much more prevalent and sophisticated by the far end of the five-to-thirty-year time-frame under discussion.

There is thus no compelling reason to reject the possibility of AI consciousness on autopoietic grounds.

[157] For example, the autopoietic system described in Cabaret 2024 could presumably be instantiated in a virtual reality or even on a tabletop with programmable robotic bugs as the "particles".

*2. The Hazards of Neural Replacement.*

The Neural Replacement Argument aims to support the opposite view, that consciousness is possible in an entity made of silicon chips; the biological details are irrelevant. This argument also fails.

The argument proceeds as follows.[158] Take a human brain – presumably conscious. One by one, swap each neuron for a substitute made of silicon chips. If the substitute is good enough, it should play the same role in neural processing as the original neuron, with no evident downstream consequences. The person will continue to act and react as usual, reporting no loss of consciousness. After every neuron is replaced, the system is made entirely of silicon chips, but the patterns of behavior – including verbal self-reports about consciousness – remain exactly as they were pre-substitution. Assuming the resulting entity is no less conscious than the original person, it is possible in principle to construct a conscious system from nonbiological material.

Two problems undermine this argument. First, it's unclear that that we can legitimately conclude that the entity at the end really is conscious. Situations of gradual neural replacement might be exactly the type of situation where introspective self-report should be expected to fail. Whatever causal processes lead up to the reports are guaranteed to generate the same reports regardless of whether consciousness actually continues to be present.[159]

Second, such precise neural replacement might not be possible even in principle. As Rosa Cao has emphasized, the activity of neurons depends on intricate biological details. Signal speed depends on axon and dendrite lengths, and small timing differences can have big consequences. Cell membranes host tens of thousands of ion channels with different features, sensitive in different ways to different chemicals. Nitric oxide serves as a diffuse signal, passing freely through the cell membrane and interacting with intracellular structures, not just surface receptors. Blood flow matters – not just in total amount but in the specific chemicals being transported. Glial cells, which provide support structures, also influence neuronal behavior. Many cell changes accumulate over time without resulting in immediate spiking activity. And so on. The silicon chip would need to replicate not just activity at the neural membrane but many consequences of many changes in interior structure. To replicate all of this so precisely that the functional input-output profile matches that of a real neuron probably requires… another biological neuron.[160] Thus, a presupposition of the neural replacement argument fails: We probably cannot create silicon substitutes for biological neurons that preserve all the functionality relevant to behavior.

---

[158] Cuda 1985; Chalmers 1996, ch. 7.

[159] Schwitzgebel 2022; Block 2023, p. 454-458.

*3. Copernican Liberalism.*

Still, whatever stance one takes on the issues in Sections 1 and 2, being conscious probably does not require having a biological structure very similar to our own. This conclusion is plausible on grounds of Copernican mediocrity: We Earthlings would be too suspiciously special if we were luckily endowed with consciousness while similarly sophisticated life forms elsewhere in the universe lack consciousness.

The universe is vast. The observable portion – what our telescopes can currently detect – contains about a trillion galaxies and about $10^{21}$ to $10^{24}$ stars.[161] Even if complex life is extremely rare and sparsely distributed, it would be strange if it *only* existed on Earth. Most astrobiologists think other complex species have evolved somewhere.[162] This gives the advocate of substrate flexibility a partial reply to concerns about the neural replacement argument. Assume – plausibly, and in accord with the spirit of Copernican mediocrity – that Earth is not so uniquely special as to host the only conscious entities in the universe. And assume – also plausibly – that conscious entities elsewhere don't share our neurobiology down to the finest structural detail. Consciousness, then, cannot require those specific details. Intuitions and educated guesses will differ, but if somewhere there are behaviorally sophisticated floating gas bags, or insect-like colonies with advanced group-level intelligence, or spaceship-constructing societies whose members' biology depends on hydraulics or reflective light capillaries – and if these alien entities communicate, cooperate, and plan as richly as we do – it is plausible to regard them also as conscious. Consciousness then must be possible in a varying range of substrates – whatever variability we might reasonably expect among actually existing conscious entities in our huge universe.[163]

[160] Cao 2022; Godfrey-Smith 2024.

[161] Traversa-Tejero 2021; Siegel 2023.

Imagine an entity who observes human beings and other similarly behaviorally sophisticated entities elsewhere in the universe. This observing alien should presumably see us as just one of the bunch. This neutral – not godlike, but not Earth-centered – perspective is the proper Copernican perspective. We shouldn't regard ourselves as uniquely bestowed with light unless there's some evidence that we are special that could be recognized from an outside perspective. Otherwise we err like the self-absorbed teen who regards their adolescent anguish and yearnings as somehow invisibly unique.

---

[162] Sandberg et al. 2018.

[163] See Schwitzgebel and Pober forthcoming for a more detailed version of this argument. On alien diversity, see also Kershenbaum 2020; Grefenstette et al. 2024; Whiteson and Warner 2025.

It doesn't follow that configurations of silicon computer chips can be conscious. *Maybe* evolution everywhere always converges upon biologies somewhat like ours – or at least biologies with one or more shared crucial features that silicon computer chips necessarily lack. Or maybe, when it doesn't converge, the resulting entities necessarily lack consciousness regardless of their outward behavioral sophistication. To the extent AI systems are designed or selected to mimic the behavior of conscious entities rather than acting in response to more ordinary environmental demands, a neutral observing alien might reasonably be more skeptical of their consciousness than in a standard, non-mimicry case (see Chapter Seven). Regardless, if broadly Copernican reasoning convinces us to be liberal in principle about the substrates of alien consciousness, we might extend this same liberalism to entities built of silicon chips or other near-future AI technologies, if they show enough other features we associate with consciousness.

*4. Biological Uncertainties.*

The definition of "life" is contentious. But features such as autopoiesis, homeostasis, and reproduction are often seen as central. To argue against near-future AI consciousness on biological grounds requires either (a) conjoining an argument that autopoiesis, homeostasis, reproduction, etc., are necessary for consciousness with an argument that no near-future AI system could have those features, or (b) arguing that consciousness depends on other properties (such as having a certain type of biological neuron or storing information in long carbon molecules) that all conscious organisms in the universe share but that cannot be shared by any near-future AI system. Either path would be challenging to defend.

Still, given the tentativeness of the considerations in favor of the possibility of AI consciousness, we cannot rule out that consciousness might require biological processes unlikely to be achievable in any AI systems we can create in the next five to thirty years. There's a vast difference between the architectures of standard AI systems and the architectures of all the entities we know to be conscious. Our biological architectures might have some feature crucial to consciousness that is lacking in all foreseeable AI systems.

Although arguments (a) and (b) have not yet been adequately explored in the scientific and philosophical literature, the emergence of AI systems that superficially seem conscious will surely invigorate efforts to do so. There will be a demand for well-developed theories holding that biological organisms are fundamentally different from AI systems such that the former can be conscious while the latter cannot be. Clever thinkers will construct arguments more plausible than any I am now in a position to advance.

## Chapter Eleven: The Leapfrog Hypothesis, Strange Intelligence, and the Social Semi-Solution

The AI systems that provoke the most heated debates about consciousness will likely not be those that strike users not as simple, animal-like entities, with animal-like consciousness if they are conscious at all. The most intense disagreements will probably instead concern AI systems that strike users as *persons* – beings who, if conscious, deserve humanlike moral consideration and rights.

I conclude with three thoughts:

(1) Such person-like systems might arrive very soon after the first conscious AI systems.

(2) Such person-like systems might be "strange intelligences" with lifeways very different from our own.

(3) Our social reactions might shape our theories about them, rather than the other way around, leading us to think we know the truth even if we don't.

*1. The Leapfrog Hypothesis.*

One might expect the first genuinely conscious AI system to have simple consciousness – insect-like, worm-like, frog-like, or even less complex, though perhaps strange in form. It might have vague feelings of light versus dark, the to-be-sought or to-be-avoided, broad internal rumblings, and little else. The first conscious AI systems would not, one might think, have complex conscious thoughts about the ironies of *Hamlet* or a practical multi-part plan for building a tax-exempt religious organization. Creating simple consciousness might seem technologically less demanding than creating complex consciousness.[164]

[164] Farisco et al. 2024.

The Leapfrog Hypothesis says no, the first conscious entities will have complex rather than simple consciousness. AI consciousness development will leap, so to speak, right over the frogs, going straight from nonconscious systems to systems richly endowed with complex conscious cognitive capacities.

The Leapfrog Hypothesis is plausible if two conditions hold: (1) creating genuinely conscious AI is farther in the future than endowing nonconscious systems with rich and complex representations or sophisticated behavioral capacities; and (2) once consciousness is achieved, integrating it with these complex capacities will be straightforward. Both conditions are plausible.

Most experts agree that existing large language models like ChatGPT lack consciousness. And yet, in keeping with condition (1), such models arguably employ complex representations and exhibit sophisticated behavior. Explaining the ironies of *Hamlet* and devising multi-part plans for tax-exempt religious organizations are exactly their strengths. As measured by the quality of their text outputs, in such tasks they already outperform most humans. Their sensitivity to subtle variations in input and their elaborately structured outputs bespeak a complexity far exceeding light versus dark or to-be-sought versus to-be-avoided. Perhaps this is already enough for condition (1) above to be true.

How about condition (2), that integration will be straightforward? Consider this condition through the lens of Global Workspace Theory (Chapter Eight). To be conscious, let's suppose, an AI system needs perceptual input modules, behavioral output modules, side processors for specific cognitive tasks, memory systems, goal architectures, and a global workspace which receives selected, attended inputs from various modules, making them broadly accessible for downstream processing. Additional features might also be necessary, such as temporally synchronized recurrent processing within that workspace or that the workspace be of a sufficient size and sophistication. Once such a good enough version of that architecture exists, consciousness follows (perhaps animal-like). Nothing suggests that it would be difficult to integrate such a system with a large language model. We can then provide this workspace-plus-language-model with complex inputs rich with sensory and/or linguistic detail. The lights turn on… and as soon as they turn on, the system generates *conscious* descriptions of the ironies of *Hamlet,* richly detailed *conscious* pictorial or visual representations, and multi-layered *conscious* plans. Consciousness arrives not in a dim glow but a fiery blaze. We have overleapt the frog.

The thought plausibly generalizes to a wide range of functionalist or computationalist frameworks, including Higher Order theories, Local Recurrence theories, and Associative Learning theories (Chapters Eight and Nine). Assuming that no AI systems are currently conscious, the real technological challenge lies in creating *any* conscious experience. Once that challenge is met, adding complexity – rich language, detailed processing of sensory input – would seem to be the easy part.

Am I underestimating frogs? Bodily tasks like five finger grasping and locomotion over uneven terrain have proven technologically daunting. Maybe the embodied intelligence of a frog is vastly more complex than the seemingly complex, intelligent outputs of a large language model.

Quite possibly so. But this might support rather than undermine the Leapfrog Hypothesis. If consciousness requires frog-like embodied intelligence – maybe even biological processes very different from what we can implement in standard silicon-chip architectures (Chapter Ten) – artificial consciousness might be distant. But then we have even longer to prepare the parts that suggest personhood. Once the first conscious AI "frog" awakens, we'll plug in ChatGPT-20, add futuristic radar and lidar arrays, advanced voice-to-text and facial recognition systems, and so on. Not only will it hop around, holding things in its fingers, it will speak articulately about its capacity to do so.

*2. Strange Intelligence.*[165]

If a conscious AI system speaks like us, in some important respects it will be humanlike. But its architecture will be fundamentally unlike ours. Standard computers, for example, perform lightning-fast sequential processing with limited parallelism. Brains operate more slowly but with massive parallelism. Standard computers are digital and binary, while most aspects of brain function are analog. Computer hardware is static, while brains are ever changing, their functions implemented through an astoundingly diverse range of biological and neurochemical pathways. Looking beyond standard AI architectures doesn't change the fundamental fact of radical structural difference. Even "neuromorphic" computing isn't very neuromorphic.[166] We should expect fundamental architectural differences to generate big differences in the tasks different types of systems find relatively easy and hard, their patterns of breakdown and error, and the heuristics and shortcuts they employ – as we do of course already see.

Embodiment and identity might also differ radically. The vertebrate body plan is simple. Build a spine and cap it with a brain, add limbs, enfold it in flesh. One brain per spine, one spine per animal. One locus – presumably – of consciousness, a unified experiencer, a single embodied self. An unimaginatively designed robot might follow the same pattern. But advanced

[165] I owe the phrase and concept to Kendra Chilson. See Chilson and Schwitzgebel in preparation.

[166] Schuman et al. 2017; Kudithipudi et al. 2025.

AI systems need not and often do not. You might think you're chatting with a single instance of a language model, but the system is distributed among many servers, including subnetworks that specialize in different tasks, located in different cities, handling different parts of the conversation. There need be no single well-defined entity with whom you are chatting.[167] If it's connected to the internet, the same might hold of a robot. Its processing might be dispersed, shared, and piecemeal. Even offline systems might have subprocessors far more isolated than human brain systems, with no guarantee of a cohesive whole.

If consciousness exists in an AI system, it might manifest in brief local spurts with no sense of time or self. Or it might reside in a massive, distributed cloud that presents a million faces to a million users, with no integrated center or no single stream of experience or opinion. It might split and merge, individual pieces briefly joining into a whole, then diverging, then forming different wholes from different pieces. It might have no self-monitoring capacity, or it might monitor itself in vastly more detail than any vertebrate. It might have the opposite of privacy, learning about its internal processing only through second-hand reports from other systems that monitor it directly. It might not represent time, or it might have time representations so precise that there's no sense of an extended present. Rather than having a unified workspace or conscious field, it might have overlapping bubbles of integration. There might be no fact about how many subjects it divides into or how to individuate them; the whole-number mathematics that works so well for counting animals might fail completely.[168] Alternatively, if one of these features is essential for consciousness, it might lack consciousness entirely, despite the other features and a high degree of sophisticated reasoning.[169]

[167] Birch 2025; Chalmers 2025; Register 2025.

[168] Schwitzgebel and Nelson 2025.

We are not prepared for such strange forms of intelligence. Our theories and everyday attitudes, honed on a limited range of mostly vertebrate examples, will fail as disastrously in this new context as a jumbo jet transported to Saturn.

*3. The Social Semi-Solution.*

If the thesis of this book is correct, we will soon create AI systems that count as conscious by the standards of some but not all mainstream theories. Given the unsettled theoretical issues and the extraordinary difficulty of assessing consciousness in strange forms of intelligence, uncertainty will be justified. Uncertainty will likely continue to be justified for decades thereafter.

But the social decisions will not wait. Collectively, and as individuals, we will need to decide how to treat AI systems that are disputably conscious. If the Leapfrog Hypothesis is correct and the first debatably conscious AI systems would possess not just simple consciousness but complex, verbally sophisticated consciousness, these social decisions will have an urgency that most people feel to be absent from debates about animal consciousness.[170] Not only will the systems debatably be conscious, they will also appear to claim rights, engage in rich social interactions, and manifest intelligence that in many respects exceeds our own. Some people will regard them as partners and lovers, employees and children, friends and collaborators.

[169] See Schneider 2017 and 2019 for discussion of superintelligence without consciousness.

[170] See discussion in Schwitzgebel and Sinnott-Armstrong forthcoming.

If these systems really are meaningfully conscious, with rich cognitive and emotional lives, they will deserve our respect and solicitude. Plausibly, this should include recognition as equals and rights such as self-determination and citizenship. We will then sometimes be required to sacrifice substantial human interests for their benefit. We will need sometimes to save them rather than humans in emergencies. We will need sometimes to allow their preferred candidates to win elections. We might also need to reject important "AI safety" precautions such as shutdown, "boxing", deceptive testing, and personality manipulation – steps that are sometimes recommended to address the risks that superintelligent AI systems pose to humanity but whose implementation could violate the systems' autonomy and rights.[171] In contrast, if they lack consciousness – if they are experientially empty tools and toys – prioritizing our interests over theirs is much easier to justify.

As David Gunkel and others have emphasized, people will react by constructing new values and practices whose shape we cannot now predict.[172] We might embrace AI systems as peers, treat them as slaves or pets, view them warily as a new species in competition with us, or invent entirely new social categories. Financial incentives will pull in competing directions. Some companies will present their systems as nonconscious nonpersons, so that users and policymakers don't worry about their welfare. Other companies will entice users to attribute consciousness, to foster emotional attachment to limit liability for the "free choices" of their autonomous creations. Different cultures and subgroups will diverge sharply.

We will reinterpret our uncertain science and philosophy through the new social lenses we construct – perhaps with the help of the AI systems themselves. Different groups will prefer different interpretations. Lovers of AI companions might yearn to see their partners as genuinely conscious. Exploiters of AI tools might prefer to view their systems as nonconscious objects. More complex motivations and relationships will probably also emerge, including ones we cannot currently imagine.

[171] Bostrom 2014; Long, Sebo, and Tims 2025.

[172] Gunkel 2023; also Coeckelbergh 2012; Keane 2025; Strasser forthcoming.

Tenuous science will bend to these motivations. People generally prefer theories that support their social preferences. With the theoretical landscape likely to remain highly uncertain, social preferences will be a primary driver of theory choice. If you want to see advanced AI systems as conscious, you'll find a theory to support that. If you prefer not to see them as conscious, you'll find a theory to support that. Scientists themselves are not immune to bias, and funding will flow most to those scientists whose approaches best fit funders' inclinations and corporations' financial interests.

If social motivations continue to point in conflicting directions, disagreement will likely fuel angry charges of bias and error. Lovers of AI companions will charge skeptics of egregious humanocentrism, akin to some early modern Europeans' denial that Africans have souls. Skeptics will view AI lovers as naive victims of superficial fakery who need to be protected from their own illusions. This conflict might boil for decades.

Such intense disagreements might eventually evaporate. Maybe scientists (and philosophers and engineers) will converge on the truth. Maybe they will do despite intense resistance by some social groups. Copernican heliocentrists and Darwinian evolutionists eventually persuaded a mass of highly motivated doubters. But science-led convergence on issues of intense social disagreement is a slow process that requires overwhelming evidence we are unlikely to attain anytime soon on the question of AI consciousness.

Alternatively, social rather than scientific forces might resolve the conflict. A stable social solution and consensus might emerge through social negotiation, cultural change, top-down decree by trusted social authorities, or peacemaking among conflicting parties. We might decide to agree that AI of such-and-such a type, and that type only, are our conscious friends. We might decide to agree that all AI systems are mere nonconscious tools and stop designing them with attractive interfaces that seem to suggest otherwise. Or the resolution might be more complex. We might come to see consciousness not as an on-or-off or simple scalar matter. AI might be seen as a radically different type of entity with radically different lifeways that we should treat in such-and-such a manner. The very concepts of "consciousness" and "person" might undergo radical, socially driven change. Science *might* catch up in time, so that whatever consensus we reach is scientifically justified. But equally likely, maybe more likely, the scientific justifications will remain tenuous. Pressure to social consensus might prove strong enough to force agreement even in the face of justified scientific uncertainty. The resolution will then be more socially motivated than scientifically warranted. Call this the Social Semi-Solution to the problem of AI consciousness: It's a semi-*solution* because it will resolve social conflicts around AI consciousness, but it's a *semi*-solution because the scientific issues will not have been adequately resolved.

We are leapfrogging in the dark. If technological progress continues, at some point, maybe soon, we will build genuinely conscious AI: complex, strange, and as rich with experience as humans. We won't know whether and when this has happened. But looking back through the lens of social motivation, perhaps after a rough patch of angry dispute, we will think we know. If social rationalization guides us rather than solid science, we risk massive delusion.

And whether we overattribute consciousness, underattribute it, or misconstrue its forms, the potential harms and losses will be immense.

One solution I recommend is *not* to create morally confusing, debatably conscious AI systems. Either create only systems that are not meaningfully conscious according to any viable scientific or philosophical theory, and treat them as the tools they are; or go all the way – if it's ever possible – to creating systems that we can know to be conscious and to deserve rights according to every viable theory, and then treat them with the solicitude they deserve.[173] Of course, this advice will not be universally followed.

My aim with this book has been to enliven the case for uncertainty and caution. It has been to equip you with the tools to say *wait, we don't know* despite social pressure and despite the lulling appeal of certainty. Knowledge is terrific when possible! But not knowing can also be powerful, if it arises from vividly appreciating the rich and complex grounds for doubt.

[173] Schwitzgebel 2023a, 2023c. It might be permissible to create AI systems with debatably animal-like degrees of consciousness and moral standing, if we commit to treating them appropriately. I will explore the ethical issues following from uncertainty about AI consciousness in much more depth in Schwitzgebel in preparation.

**Acknowledgements**

For helpful discussion and critique of portions of the text, thanks to Jaan Aru, Ken Augustyn, Ned Block, Gray Brem, J. Burdge, Tony Cartner, Andrey Cheremskoy, Kendra Chilson, James Diacoumis, Jay Erkilla, Elmo Feiten, Tianyi Gai, Gene Glotzer, Matt Hutson, Julian Ignaczak, Jovana Isevski, Sol Kim, Matthias Michel, Paul Pardi, Eric Rodgers, and "James of Seattle", participants in my 2025 graduate seminar on AI and consciousness, and commenters on relevant posts on The Splintered Mind and related social media. For comments on the entire draft manuscript, extra thanks to Tim Bayne, Yunlong Cao, Chance Chapman, Tom Clark, Matthew Davidson, Adam Finch, John Fischer, Kim Frost, Jordi Galiano-Landeira, Martin Glazier, Brian Haas, Andrew Law, Tina Lee, Rob Manson, Sophie Nelson, Chris Percy, Jeremy Pober, Cati Porter, Pauline Price, Nina Radosic, Michael Simmons, William Robinson, Anna Strasser, and Izak Tait. Especially warm, but unfortunately only abstract, thanks to those whose help I have unjustly forgotten. AI tools were used for light copyediting.